\documentclass[10pt]{article} 

\usepackage{algorithm}
\usepackage{algorithmic}

\usepackage[accepted]{tmlr}

\usepackage{amsmath,amsfonts,bm}

\def\eqref#1{equation~\ref{#1}}

\def\1{\bm{1}}

\DeclareMathAlphabet{\mathsfit}{\encodingdefault}{\sfdefault}{m}{sl}
\SetMathAlphabet{\mathsfit}{bold}{\encodingdefault}{\sfdefault}{bx}{n}

\usepackage{url}
\usepackage{graphicx}
\usepackage{booktabs}
\usepackage{tabularx}
\usepackage{tcolorbox}
\usepackage{enumitem}
\usepackage{wrapfig}

\definecolor{codegreen}{rgb}{0,0.6,0}
\definecolor{codegray}{rgb}{0.5,0.5,0.5}
\definecolor{codepink}{RGB}{252, 142, 172}
\definecolor{codepurple}{rgb}{0.58,0,0.82}
\definecolor{backcolour}{RGB}{245,245,245}
\usepackage{listings}

\lstdefinestyle{marx_style}{
    backgroundcolor=\color{backcolour},   
    commentstyle=\color{magenta},
    keywordstyle=\color{blue},
    numberstyle=\tiny\color{codegray},
    stringstyle=\color{codepurple},
    basicstyle=\fontfamily{\ttdefault}\footnotesize,
    breakatwhitespace=false,        
    breaklines=true,                
    keepspaces=true,    
    frame=single,
    numbersep=5pt,                  
    showspaces=false,              
    showstringspaces=false,
    showtabs=false,               
    tabsize=2,
    classoffset=1, 
    keywordstyle=\color{violet},
    classoffset=0,
    language=Python
}
\definecolor{mydarkred}{rgb}{0.6,0,0}
\definecolor{mydarkgreen}{rgb}{0,0.6,0}
\usepackage[colorlinks,linkcolor=mydarkred,citecolor=mydarkgreen]{hyperref}

\title{LLM-Based World Models Can Make Decisions Solely, \\ But Rigorous Evaluations are Needed}

\author{\name Chang Yang$^1$, Xinrun Wang$^2$\thanks{Corresponding author}\ , Junzhe Jiang$^1$, Qinggang Zhang$^1$, Xiao Huang$^1$ \\\\
\addr $^1$The Hong Kong Polytechnic University, $^2$Singapore Management University \\\\
\email chang.yang@connect.polyu.hk, xrwang@smu.edu.sg
}

\def\month{03}  
\def\year{2026} 
\def\openreview{\url{https://openreview.net/forum?id=XmYCERErcD}} 

\begin{document}

\maketitle

\begin{abstract}
World model emerges as a key module in decision making, where MuZero and Dreamer achieve remarkable successes in complex tasks. Recent work leverages Large Language Models (LLMs) as general world simulators to simulate the dynamics of the world due to their generalizability. LLMs also serve as the world model for deliberative reasoning in Reasoning via Planning (RAP) and Tree of Thought (ToT). However, the world model is either evaluated as a general world simulator, or as a functional module of the agent, i.e., predicting the transitions to assist the planning. 
This paper argues that \textbf{LLM-based world models can make decisions solely, but rigorous evaluations are needed}. We first present the two key observations to showcase how LLM-based world models can make decisions solely, and then present the three key observations to demonstrate why current evaluation framework of LLM-based world models is not sufficient. Then, we present our suggested evaluation framework: \emph{policy verification}, \emph{action proposal}, and \emph{policy planning}, where the world model is used for decision making solely, and finally we leverage the \textbf{31} diverse environments from~\citep{wang2023bytesized,wang2024language} and curate the rule-based policy of each environment for diverse evaluations. 
The key findings include: i) GPT-4o significantly outperforms GPT-4o-mini on the three main tasks, especially for the tasks which require the domain knowledge, e.g., scientific tasks, ii) the performance of the LLM-based world models depends predominantly on their performance in key steps, while the total number of steps required for task completion is a weak indicator of task difficulty with critical bottleneck steps playing a more decisive role and iii) the combination of world models' functionalities for decision making tends to increase performance instability in our experiments, which can partially obscure the performance gap between stronger and weaker model.
\end{abstract}

\section{Introduction}

The remarkable achievements of MuZero~\citep{schrittwieser2020mastering} and Dreamer~\citep{hafner2019dream,hafner2021mastering,hafner2025mastering,hafner2025training} have established world models~\citep{ha2018world} as fundamental components in decision-making systems. World models serve as learned simulators that encode rich representations of environment dynamics, enabling agents to predict future states conditioned on their actions. Rather than learning direct policy mappings from observations to actions, world models learn the underlying transition dynamics $T(s, a) \rightarrow s'$ and reward functions $R(s, a)$, providing a compressed yet informative representation of how the environment evolves. These models have demonstrated effectiveness across several key areas with concrete achievements: i) \textit{Generalization to novel tasks.} World models enable transfer learning by capturing reusable environment dynamics~\citep{byravan2020imagined}. For instance, agents trained with world models can adapt to new task objectives without relearning dynamics~\citep{robey2021model}, and demonstrate robust performance across distribution shifts in robotics applications~\citep{young2023benefits}. This capability is particularly valuable when the cost of re-training from scratch is prohibitive. ii) \textit{Efficient planning.} By simulating possible futures in the learned model, agents can evaluate action sequences through "mental simulation" rather than costly real-world interaction~\citep{sekar2020planning,hamrick2021role}. MuZero leverages this capability to achieve superhuman performance in Go, chess, and Atari games by planning over imagined trajectories~\citep{schrittwieser2020mastering}. The ability to explore hypothetical scenarios without environmental interaction dramatically improves sample efficiency and enables safer exploration in high-risk domains. iii) \textit{Offline learning from fixed datasets.} World models have proven especially valuable in offline reinforcement learning settings~\citep{schrittwieser2021online,yu2020mopo,yu2021combo}, where agents must learn purely from pre-collected datasets without additional environment interaction. By learning accurate dynamics models from historical data, agents can perform policy optimization in the model space, generating synthetic rollouts to augment limited real data. This capability opens possibilities for training agents in domains where online interaction is impractical or expensive, such as healthcare and autonomous driving.

Recent developments have extended world models beyond traditional reinforcement learning to interactive simulation systems, demonstrating their potential as general-purpose world understanding engines. Notable examples include the Genie series~\citep{bruce2024genie,deepmind2024genie2,deepmind2025genie3}, which learns controllable world models from unlabeled video data and enables users to interact with generated game-like environments through action inputs. Vista~\citep{gao2024vista} extends this paradigm to autonomous driving, learning world models that can simulate diverse driving scenarios and enable policy training in purely synthetic environments. Additional systems such as MatrixGame~\citep{zhang2025matrix,he2025matrix}, which generates playable game environments from natural language descriptions, and Hunyuan-GameCraft~\citep{li2025hunyuan}, which creates interactive 3D game worlds, further illustrate the expanding capabilities of world models in interactive domains. These systems share a common theme: they learn predictive models of complex, high-dimensional dynamics (visual observations, physics, agent interactions) and expose these models through interactive interfaces. These advances point toward world models serving as foundational components for artificial general intelligence systems with grounded understanding of world dynamics, moving beyond narrow task-specific applications to general-purpose simulation and interaction.

Large Language Models (LLMs) have emerged as a promising approach to world modeling, leveraging their broad knowledge acquired from large-scale text corpora. Building on their success in natural language tasks~\citep{brown2020language,openai2023gpt4}, recent works explore LLMs as general world models that provide environment knowledge for complex reasoning and decision-making tasks. The key hypothesis is that LLMs, having been exposed to vast amounts of text describing how the world works, have internalized implicit models of common-sense physics, social dynamics, and domain-specific knowledge (e.g., mathematics, science, games). Through fine-tuning on environment interaction data, i.e., sequences of states, actions, and outcomes, LLMs can predict action sequences across diverse tasks while maintaining their broad capabilities~\citep{xiang2023language,wang2024language,xie2024making}. For example, an LLM fine-tuned on household task demonstrations can predict valid action sequences for cooking or cleaning tasks while retaining its language understanding and reasoning abilities. LLMs serve as world models in both explicit and implicit ways, depending on how the state prediction capability is utilized: i) \textit{Explicit world modeling.} In approaches like Reasoning via Planning (RAP)~\citep{hao2023reasoning} and Reason for Future, Act for Now (RAFA)~\citep{liu2023reason}, LLMs directly predict next states given current states and actions, functioning as explicit transition models $T_{\text{LLM}}(s_t, a_t) \rightarrow s_{t+1}$. For instance, in BlocksWorld~\citep{valmeekam2023planbench}, given a state description like ``Block A is on Block B, Block C is on the table'' and an action ``move Block A to Block C,'' the LLM predicts the resulting state: ``Block A is on Block C, Block B is on the table, Block C is on the table.'' These predicted states are then used by planning algorithms (e.g., Monte Carlo Tree Search, breadth-first search) to evaluate action sequences and select optimal plans. The explicit separation allows for modular analysis of world model accuracy independent of planning algorithm performance.
ii) \textit{Implicit world modeling.} In widely-used approaches like Tree of Thoughts (ToT)~\citep{yao2023tree} and Graph of Thoughts (GoT)~\citep{besta2024graph}, LLMs perform world modeling implicitly as part of their reasoning process. Rather than explicitly outputting state descriptions, these methods have LLMs generate and evaluate intermediate reasoning steps (``thoughts'') that implicitly represent states in a problem-solving trajectory. For example, when solving a math problem, each thought represents a partial solution state, and the LLM must implicitly model how different reasoning steps transition between these states. The LLM evaluates the promise of each thought (analogous to value estimation in RL) to guide search through the solution space. We provide a comprehensive review of related work in Section~\ref{app:related_work}.

Despite these celebrating progress, the role of LLM-based world models in decision models is still unclear for researchers. In this paper, we present a comprehensive view: \textbf{LLM-based world models have the potential to make decisions solely through the combination of policy verification and action proposal capabilities, but rigorous evaluation frameworks are needed to assess this capability.} We first present the two key observations to showcase how LLM-based world models can make decisions, and then present the three key observations to demonstrate why current evaluation framework of LLM-based world models is not sufficient. Then, we present our suggested evaluation framework: i) \textbf{policy verification}: verifying whether the policy can complete the task, ii) \textbf{action proposal}: proposing the top-$K$ actions that can potentially complete the task, and iii) \textbf{policy planning}: finding the policy solely with the combination of the different functionalities, i.e., policy verification and action proposal. Finally, we leverage \textbf{31 diverse environments} from~\citep{wang2023bytesized,wang2024language} with different tasks varying from daily tasks, e.g., washing clothes, to scientific tasks, e.g., forging keys, and curate the rule-based policy for each environment for the evaluation and conduct the comprehensive evaluation of the advanced LLMs, i.e., GPT-4o and GPT-4o-mini, on the environments for three tasks under various settings, {which serves as a proof of concept for the suggested evaluation framework}. The \textbf{key findings} include: i) GPT-4o significantly outperforms GPT-4o-mini on the three main tasks, especially for the tasks which requires the domain knowledge, e.g., scientific tasks, ii) the performance of the LLM-based world models depends predominantly on their performance in key steps, while the total number of steps required for task completion is a weaker indicator of task difficulty and critical bottleneck steps play a more decisive role and iii) {the combination of world models' functionalities for decision making tends to increase performance instability in our experiments, which can partially obscure the performance gap between stronger and weaker model. We hope this work can encourage researchers to rethink the evaluation of world models and further advance the research of the world model field. Code can be access at: \url{https://github.com/joannacyang/WorldModel_TMLR}.

\section{Preliminaries}


\textbf{Markov Decision Process (MDP).} A decision making problem is usually represented as a Markov decision process (MDP)~\citep{sutton2018reinforcement}, defined by the tuple $(S, A, T, R, \gamma)$, where $S$ is the state space, $A$ is the action space, $T: S\times A\rightarrow S$ is the transition dynamics, which specifies the next state $s'$ given the current state $s$ and action $a$, $R: S\times A\rightarrow\mathbb{R}$ is the reward function, which specifies the agent's reward given the current state $s$ and action $a$, and $\gamma$ is the discount factor. The agent's policy is defined by $\pi_{\theta}: \mathcal{S}\times\mathcal{A}\rightarrow[0,1]$, parameterized by $\theta$, which takes the state $s$ as the input and outputs the action $a$ to be executed. The objective of the agent is to learn an optimal policy $\pi^{*}:=\arg\max_{\pi}\mathbb{E}_{\pi}\left[\sum_{t=0}^{\infty} \gamma^{t} r_{t} | s_{0}\right]$ is the expected return and $s_{0}$ is the initial state. 


\textbf{Large Language Models (LLMs).} Large Language models (LLMs) learn from text data using unsupervised learning. LLMs optimize the joint probabilities of variable-length symbol sequences as the product of conditional probabilities by $P(x)=\prod\nolimits^n_{i=1}P(s_i|s_1,...,s_{i-1})$, where $(s_1, s_2, ..., s_n)$ is the variable-length sequence of symbols. With the billions of parameters and extensive training data, the vast amounts of common knowledge encoded in LLMs lead to the remarkable generalization across various NLP tasks with simple prompting and in-context learning, without task-specific fine-tuning~\citep{touvron2023llama,openai2023gpt4}. Given the generalizability, LLMs present a promising foundation for general world models.




\textbf{LLM-based World Models.} 
The world model $\Omega$ is introduced to predict the dynamics of the environment, thus supporting the decision making process. Specifically, the world model is trained or prompted to predict the next state $s'$, the reward $r$, and the terminal function $d$, given the current state $s$ and action $a$. The world model can be one or multiple neural networks specially trained on the environments for the three prediction tasks~\citep{hafner2019dream,schrittwieser2020mastering}, which cannot generalize across different environments. Recent work leverages chain-of-thought (CoT) prompting~\citep{xie2024making}, in-context learning~\citep{wang2024language}, retrieval-augmented generation~\citep{yang2025efficient}, and fine-tuning methods~\citep{xiang2023language,lin2024learning} to transform LLMs to world models. 

\textbf{Scope of This Work.} {In this work, we intend to investigate the core roles of LLM-based world models in decision making and design the comprehensive evaluation framework. Rather than evaluating world models as general simulators or as components coupled with other modules, we propose three decoupled evaluation tasks: policy verification, action proposal, and policy planning. We empirically evaluate GPT-4o and GPT-4o-mini across 31 diverse text-based environments as a proof of concept to assess the reasonability of the suggested evaluation framework.}


\section{LLM-Based World Models Can Make Decisions Solely}

We present the two observations to illustrate how LLM-based world models can make decisions solely.

\textbf{Observation 3.1: Selecting potential actions should be an important feature for world models.} Most of the previous works in world model focus on next state and reward prediction, and the action selection is usually completed by the actors, i.e., the model trained for generate a single action for executing.  We argue that with more knowledge about the world, the world model may make a better selection of the potential actions. Besides, selecting a set of potential actions, e.g., 10 potential actions, may significantly reduce the difficulties of the tasks and improve the performance when combing with planning. World models can also be viewed as game engines~\citep{valevski2024diffusion}, which have to provide potential actions to guide fresh players to complete tasks, e.g., Red Dead Redemption 2~\citep{tan2024cradle}. Therefore, action proposal should be considered for evaluation, which can be easily implemented for the LLM-based world models.


\begin{wrapfigure}{r}{0.55\textwidth}
    \centering
    \vspace{-10pt}
    \includegraphics[width=\linewidth]{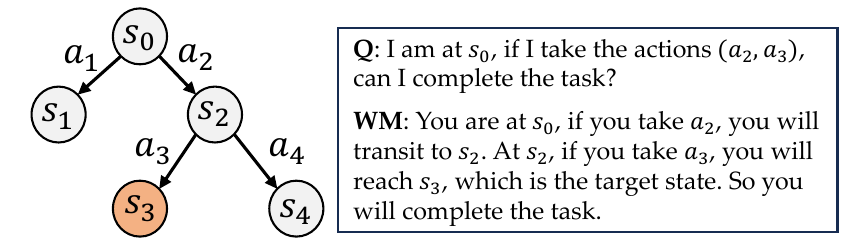}
    \vspace{-15pt}
    \caption{An example to show that the critic is not necessary. We can query all action sequences and let the world model to determine whether the action sequence can complete the task, which may be insufficient but still possible. }
    \label{fig:critic}
    \vspace{-10pt}
\end{wrapfigure}
\textbf{Observation 3.2: Planning with world models can find the policies solely.} With the prediction of the next states and the action proposal, we can leverage planning methods or search methods to find the policies. It is observed that most state-of-the-art methods for complex decision-making tasks, e.g., Chess or Go, is based on the planning with an accurate simulator~\citep{silver2018general,monroe2024mastering} or the world model~\citep{schrittwieser2020mastering}.
Most works introduce the critic (i.e., the value function) to evaluate the actions immediately for efficient planning~\citep{schrittwieser2020mastering,hao2023reasoning}. We note that the critic is not necessary for finding policies (as showed in Figure~\ref{fig:critic}) and may also influence the performance. Therefore, we argue that only the planning with the next state prediction and the action proposal is necessary when incorporating the world model in decision making. This focused approach allows for better isolation and evaluation of the world models and further highlight the importance of world models.

\section{But Rigorous Evaluations are Needed}


\label{sec:rigorous_evaluation}

\begin{wrapfigure}{r}{0.525\textwidth}
\centering
\vspace{-50pt}
\includegraphics[width=0.575\linewidth]{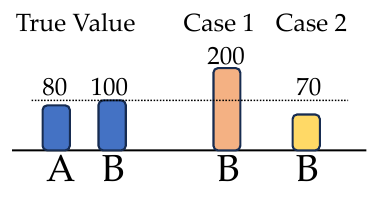}
\vspace{-25pt}
\caption{Picking the action with higher value. {The true value of A and B are 80 and 100, respectively, therefore, B is the correct action. Case 1's prediction 200 is worse than in Case 2's prediction 70 (compared to the true value 80). However, in Case 1 the most valuable action is B, which is also the case in the true value setting. This shows that more accurate predictions (Case 2) do not always lead to correct decisions (Case 1).}}
\label{fig:prediction}
\vspace{-10pt}
\end{wrapfigure}

\textbf{Observation 4.1: Prediction is important, but not that important.} An illustrative example is displayed in Figure~\ref{fig:prediction}, which indicates that more accurate predictions do not lead to correct decisions.\footnote{The issue in Figure~\ref{fig:prediction} can be elicited by various methods, e.g., rank prediction. This is just to illustrate the discrepancy between prediction and decision, motivating us to reconsider the evaluation of the world model for decision making.} This phenomenon is also observed in other decision making scenarios, e.g., financial trading~\citep{sun2023reinforcement}. The success of MuZero Unplugged~\citep{schrittwieser2021online} also demonstrate that we can learn good policies from inaccurate world models which are trained only with limited data. This motivates us that the evaluation of the world models for decision making should focus on the predictions which relevant to the desired policy, rather than as general world simulators. Besides, the decision making usually involves multiple steps and the errors of the one-step predictions are accumulated when the number of steps increases. Therefore, the accuracy of the one-step predictions is not adequate for the evaluation of the world model for decision making.

\begin{wrapfigure}{r}{0.45\textwidth}
\centering
\vspace{-10pt}
\includegraphics[width=\linewidth]{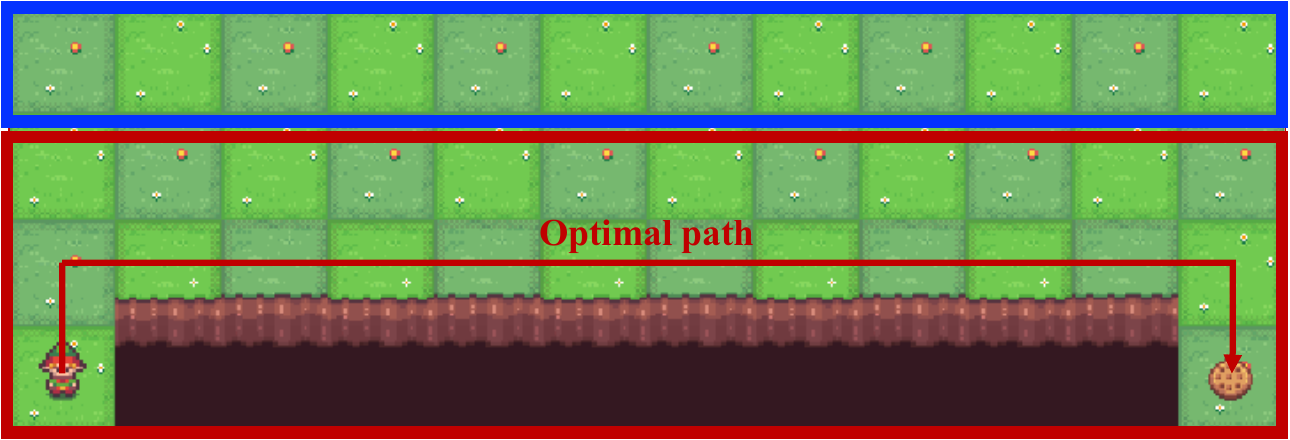}
\vspace{-15pt}
\caption{Cliff Walking~\citep{sutton2018reinforcement}. The optimal path is marked as the red arrow.}
\label{fig:state_importance}
\vspace{-5pt}
\end{wrapfigure}
\textbf{Observation 4.2: Not all state are created equal for decision making.} 
An illustrative example is displayed in Figure~\ref{fig:state_importance}, where the agent needs to go through a gridworld from start to goal while avoiding falling off a cliff. In this example, the states in the red rectangle are more important than the states in the blue rectangle, as these red states are closer to the optimal path and the blue states are rarely explored by the agent when learning to complete the tasks. This observation is supported by the fact that only a small portion of the state space will be visited ($S_{\text{visit}} \ll S$) when computing the optimal policy, e.g., AlphaZero finds the super-human policy~\citep{silver2018general} by only exploring a small proportion (less than 1\%) of the state space.
This observation suggests that world model evaluation should prioritize states that are critical for task completion.


\begin{wrapfigure}{r}{0.45\textwidth}
\centering
\vspace{-20pt}
\includegraphics[width=.65\linewidth]{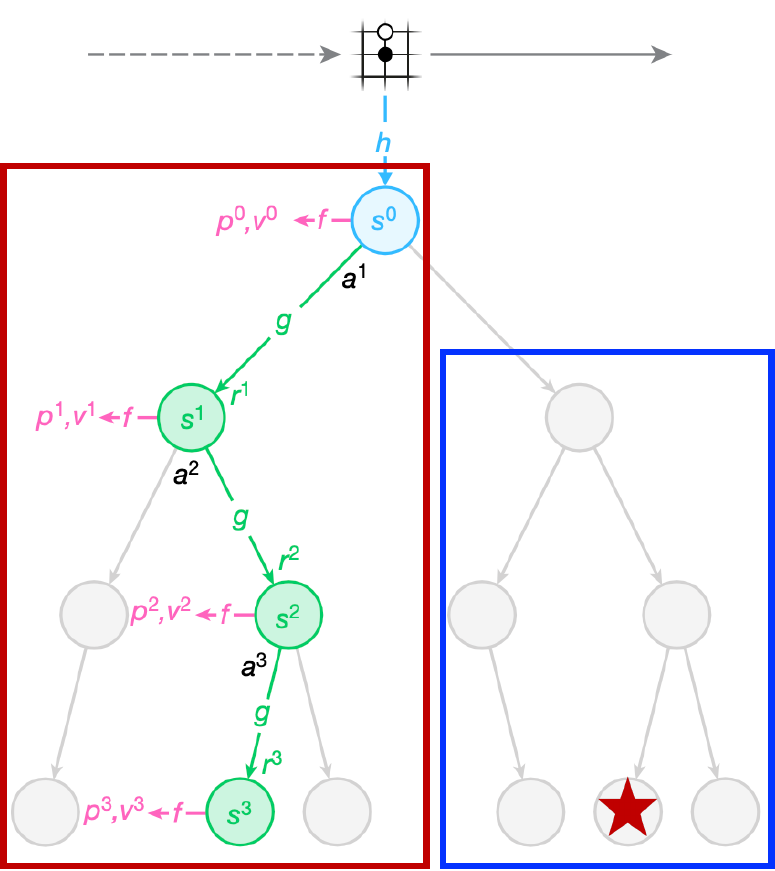}
\caption{Figure adpted from MuZero~\citep{schrittwieser2020mastering} to illustrate search process, i.e., MCTS, in model-based planing.}
\label{fig:muzero_search}
\vspace{-10pt}
\end{wrapfigure}
\textbf{Observation 4.3: Performance is usually coupled with other modules.} An illustrative example is displayed in Figure~\ref{fig:muzero_search}. The states in the red rectangle is explored by MCTS, while the optimal node (marked as the red star) is in the unexplored states in the blue rectangle. Therefore, even the world model can provide the accurate predictions, the agent's performance will be restricted by the search algorithm. In model-based planning approaches like MuZero~\citep{schrittwieser2020mastering}, the world model's performance cannot be isolated from the search algorithm, e.g., MCTS, value network, and policy network~\citep{hafner2025mastering}. When these components are tightly integrated, it becomes difficult to determine whether poor performance stems from an inaccurate world model, suboptimal planning, or ineffective value estimation. This coupling obscures our understanding of the world model's true capabilities and limitations. To address this issue, the evaluation should decouple the world models with other modules and design the tasks which specifically test the world model's functionalities.



\section{Our Suggested Evaluation Framework and Experiment Results}

\begin{wrapfigure}{r}{0.55\textwidth}
\centering
\vspace{-10pt}
\includegraphics[width=.85\linewidth]{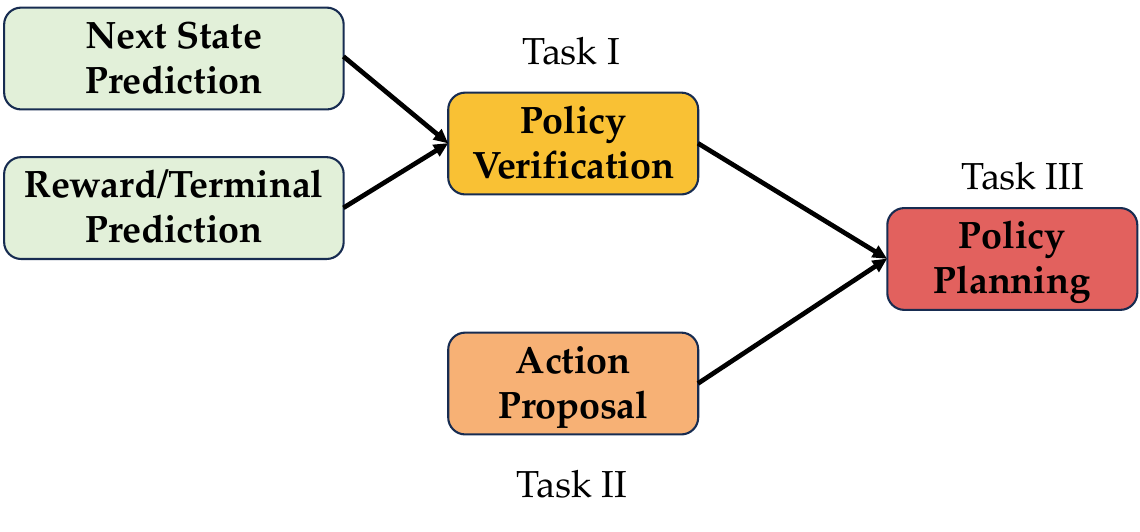}
\vspace{-13pt}
\caption{With the basic predictions, i.e., next state and reward/terminal, the \textbf{policy verification} can be completed. With \textbf{policy verification} and \textbf{action proposal} of the world model, \textbf{policy planning} can be completed.}
\label{fig:task_tax}
\vspace{-15pt}
\end{wrapfigure}
In this section, we introduce our evaluation framework and the experiment results. Specifically, we will introduce the next state prediction, the reward and terminal prediction. Then, we will introduce how the world model will be used to complete the considered three main tasks, i.e., policy verification, action proposal, and policy planning. We provide the relationship between the three tasks and the two kinds of predictions in Figure~\ref{fig:task_tax} for better understanding of the rationale behind the three tasks. We then introduce the environments and rule-based policies. Finally, we will introduce each task, the evaluation protocols, and the corresponding experimental results. Our experiments utilize GPT-4o and GPT-4o-mini as the backbone LLM architectures for the world model.\footnote{Due to the limited budget, we do not take a full list of LLMs, e.g., Claude and Gemini, into consideration.} To ensure reproducibility and minimize stochastic variance in model outputs, we configure the temperature parameter to 0. All experimental results are averaged across 30 independent trials to handle environmental stochasticity and provide statistically robust metrics. {Our evaluation results should be interpreted as diagnostic assessments of specific capabilities under controlled conditions to demonstrate the reasonability of the suggested evaluation framework, rather than measures of general world modeling ability or readiness for deployment.}



The world model considered in this work mainly follows the design in~\citep{wang2024language}, where the representation of the states includes the objects in the environments and their properties. The prompts to the LLM, e.g., GPT-4o, also include the object rules, the action rules, and the score rules, which provides the necessary knowledge of the environments for the LLM to make accurate predictions. For the \textbf{next state prediction}, we ask the LLM to predict the state changes, i.e., the change of the objects' properties, which is demonstrated to be efficient for the prediction~\citep{wang2024language}. With the predicted state changes, we can recover the full state for further predictions. For the \textbf{reward/terminal prediction}, the LLM needs to predict three features: i) gameScore: the reward received from the environment, ii) gameOver: whether the task is terminated, and iii) gameWon: whether the task is successfully completed or not. For the rules used for the prediction, we refer to~\citep{wang2024language} for more details and the code to generate the prompts is also provided in Appendix~\ref{app:next_state_and_reward_prediction} for completeness.\footnote{Due to the space constraints and the extreme length of the prompts, we cannot provide a complete example in the paper. We will open-source all the codes for readers to replicate our results.}


\textbf{Environments.} {We leverage the \textbf{31} diverse environments from~\citep{wang2023bytesized}\footnote{We note that there are 32 environments in~\citep{wang2023bytesized} and the dish-washing environment is used as the example for the world model, which is excluded for fair evaluation.} with different tasks varying from daily tasks, e.g., washing clothes, to scientific tasks, e.g., forging keys. This task suite is more related to the real physical world, including the physical objects, e.g., bulb and bathtub, and the iterations with these physical objects, i.e., turn on the hot tap to improve the temperature of the water in the bathtub. Compared with other widely used environments, such as the grid world, e.g., BabyAI~\citep{chevalier2018babyai} and the web environments, e.g., MiniWob++~\citep{shi2017world}, this task suite is more relevant to the common knowledge encoded in the LLMs. A full descriptions of the environments is in Appendix~\ref{app:env_task}.
}


\begin{figure}[ht]
    \centering
    \includegraphics[width=\linewidth]{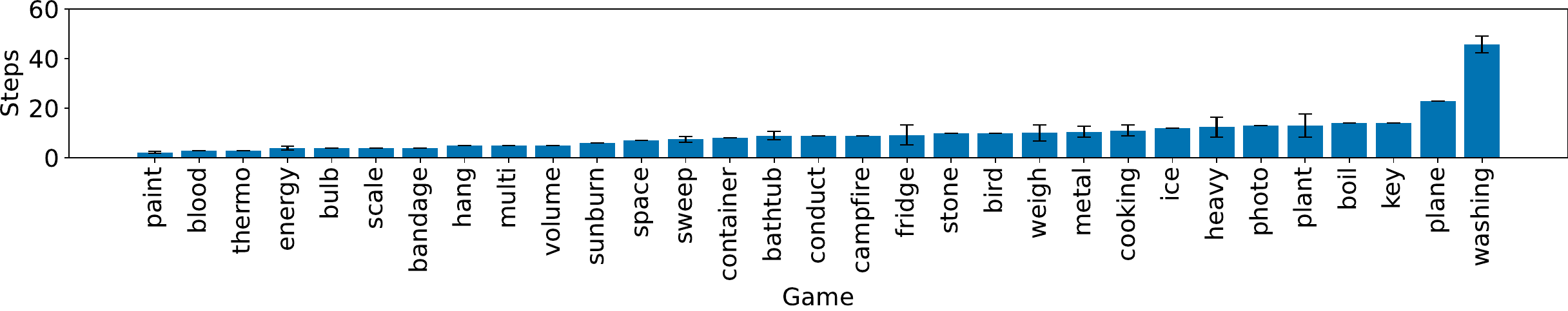}
    \vspace{-20pt}
    \caption{Number of steps to complete the tasks}
    \label{fig:steps_main}
    
\end{figure}


\textbf{Rule-based Policies.} There are various randomness in the environments, including the specific tasks, e.g., the target color can be ``orange'', ``purple'', ``green'', ``black'' in the paint task. However, for each task, only a single playthrough is provided in~\citep{wang2024language}, which is not enough for a comprehensive evaluation of the world model for decision making. Therefore, we curate the rule-based policy for each environment and verify the correctness for 200 runs. The scripts for the rule-based policies are provided in Appendix~\ref{app:rule_based_policies}, which can help readers to understand the process to complete the tasks. We provide the statics of the number of steps to complete the tasks for 200 runs in Figure~\ref{fig:steps_main}.\footnote{The names on the figure may differ from~\citep{wang2023bytesized} for plotting. Please refer to Table~\ref{tab:envs} for correspondence.}


\subsection{Task I: Policy Verification} 

\textbf{Task Description.} Motivated by \textbf{Observations 4.1 and 4.2}, we propose policy verification, one of the most straightforward tasks to evaluate the world models in terms of the multi-step predictions. The basic idea of policy verification is given an action sequence, the world model need to prediction whether the sequence can complete the task or not.
The process for the policy verification is displayed in Algorithm~\ref{alg:policy_verification}. Specifically, given the environment \texttt{env}, the action sequence $\bm{a}$ with length $N$ to verify and {$\rho$ denotes the fraction of the action sequence for which the world model is responsible for verification/planning.} We will run the game for the first $(1-\rho)\cdot N$ steps (Line~\ref{alg:policy_verification_env} of Algorithm~\ref{alg:policy_verification}), and leverage the world model to continue the last $\rho N$ steps (Line~\ref{alg:policy_verification_world_model}  of Algorithm~\ref{alg:policy_verification}). The returned results $r_{N}, d_{N}$ will be compared with the true results from the environment to evaluate the world model.

\textbf{Evaluation Protocol.} Given the action sequence $\bm{a}$ generated by the rule-based policy, we leverage the world model to verify the last $\rho$ proportion of the policy, where $\rho \in \{0.25, 0.5, 0.75, 1.0\}$. We note that when $\rho=1.0$, the world model will verify the full action sequence with only the initial observation of the environment. We say the verification of the policy is correct if all three features, i.e., gameScore, gameOver, and gameWon, are correct. {Our policy verification evaluation only tests correct (positive) policies. A complete verification capability would also need to reject incorrect (negative) policies, which we do not test. This means we measure ``false negative rate'' (incorrectly rejecting valid policies) but not ``false positive rate'' (incorrectly accepting invalid policies). True utility for planning would require low rates on both. Generating meaningful negative policies is challenging: random action sequences are trivially wrong, while plausible-but-flawed policies require domain expertise to construct. We leave comprehensive negative policy evaluation to future work, acknowledging this significantly limits claims about verification reliability for search applications. Furthermore, there would also be other action sequences to finish the tasks, where we cannot enumerate all policies to complete the tasks.}

\begin{figure*}[t]
\centering
\vspace{-20pt}
    \includegraphics[width=0.95\textwidth]{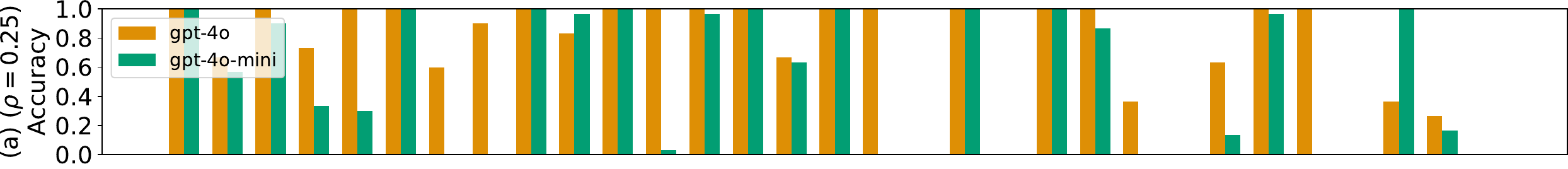}
    \includegraphics[width= 0.95\textwidth]{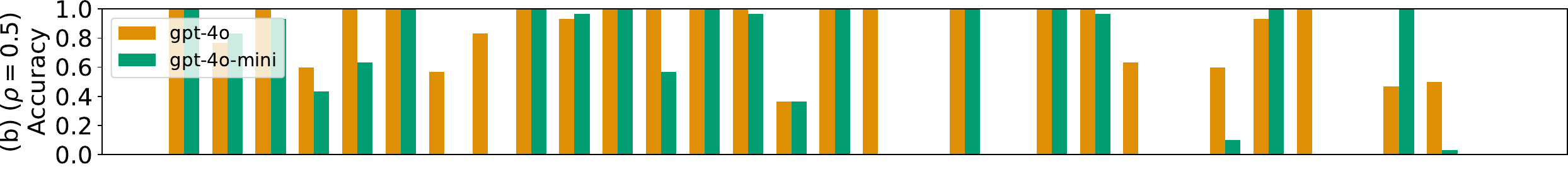}
    \includegraphics[width= 0.95\textwidth]{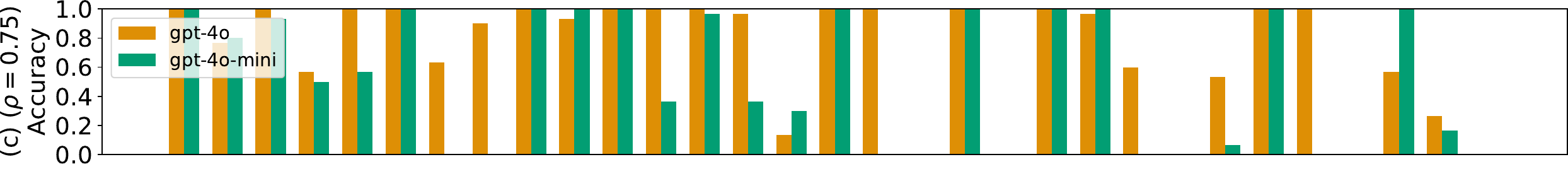}

    \includegraphics[width= 0.95\textwidth]{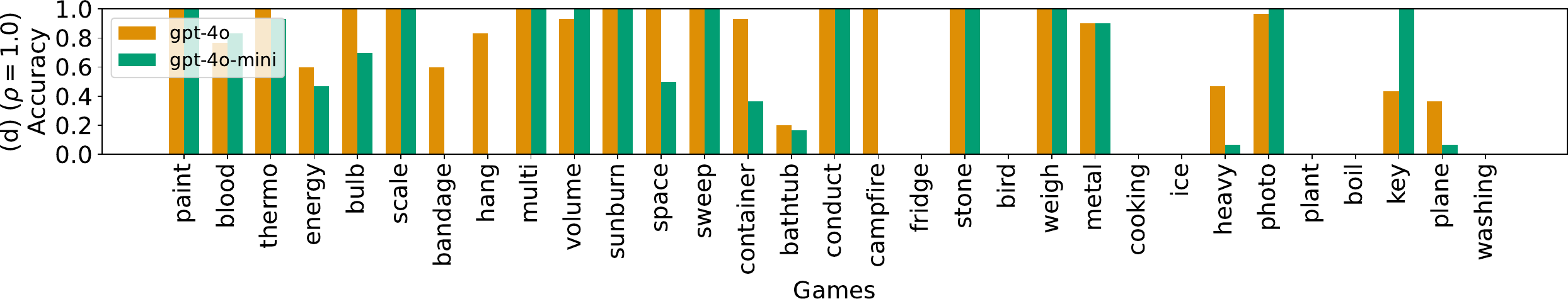}
\vspace{-10pt}
\caption{The accuracy of the world model to verify the correct policies}
\label{fig:verify}
\vspace{-10pt}
\end{figure*}

\textbf{Evaluation Results.} The policy verification results are displayed in Figure~\ref{fig:verify}.\footnote{Note that the result of the policy verification is either 0 or 1, so the error bar is not plotted.} 
We observe that GPT-4o outperform GPT-4o-mini in most tasks and especially on the tasks which requires the domain knowledge, e.g., bandage, hang, and campfire. We also observe that with more steps of the verified policies, the performance gap between GPT-4o and GPT-4o-mini is increase. With larger proportion of the action sequences to verify, i.e., $\rho$ increase, the accuracy of the verification is decreased, which indicates that the accumulation of the errors in the world model, either on the next state prediction or the reward prediction, will influence the performance of the world model. This observation is consistent to the fact that the LLM may not perform well in long-term decision making tasks. {We also observe that more steps to complete the tasks do not necessarily lead to the worse performance, which indicates that the domains of the tasks play a more important role for the policy verification, i.e., for the tasks where the LLM has enough domain knowledge, e.g., conduct, stone, weigh and photo~\citep{wang2024language}, the task would be easy even when the number of steps is large. We also provide the accuracy of the three prediction tasks separately in Appendix~\ref{app:verify}, and we found that both GPT-4o and GPT-4o-mini performs worse for predicting gameScore, while performs much better for predicting gameOver and gameWon. This indicates that the value prediction is more difficult for LLMs, which is consistent the observations from other works. During the experiments, both models frequently returned empty dictionaries, suggesting they may fail to properly follow the instructions.}

\begin{tcolorbox}[colback=blue!10, colframe=black, coltitle=black, left=0pt, right=0pt, top=0pt, bottom=0pt, arc=0.5mm, boxrule=0.5pt
]
\textbf{Takeaways}
\begin{itemize}[leftmargin=*, topsep=0pt, partopsep=0pt, parsep=0pt, itemsep=0pt]
\item Policy verification evaluates multi-step prediction accuracy by verifying whether action sequences can complete tasks, addressing the limitations of single-step prediction evaluation
\item Performance degrades with more verified steps, i.e., error accumulation in long-horizon predictions
\item GPT-4o significantly outperforms GPT-4o-mini, especially on tasks requiring domain knowledge
\item Task difficulty depends more on domain knowledge required than the total number of completion steps
\end{itemize}
\end{tcolorbox}

\begin{figure}[ht]
\centering
\begin{minipage}{0.48\textwidth}
\begin{algorithm}[H]
\caption{Policy Verification}
\label{alg:policy_verification}
\begin{algorithmic}[1]
\STATE Given the \texttt{env}, the action sequence $\bm{a}$ to verify with $N=\texttt{len}(\bm{a})$, $\rho$ the proportion of $\bm{a}$ to verify, the world model $\Omega$
\STATE $s_{0}=\texttt{env()}$\;
\FOR{$t\in\{1, 2, ..., N-1\}$}
\IF{$t<(1-\rho)\cdot N$}
\STATE $s_{t+1}, r_{t}, d_{t}=\texttt{env}(a_{t})$\label{alg:policy_verification_env}
\ELSE
\STATE {$s_{t+1}, r_{t}, d_{t}=\Omega(s_{t}, a_{t})$\label{alg:policy_verification_world_model}}
\ENDIF
\ENDFOR
\RETURN $r_{N}, d_{N}$
\end{algorithmic}
\end{algorithm}   
\end{minipage}
\hfill
\begin{minipage}{0.48\textwidth}
\begin{algorithm}[H]
\caption{Policy Planning}
\label{alg:policy_planning}
\begin{algorithmic}[1]
\STATE Given the \texttt{env}, the action sequence $\bm{a}$ with $N=\texttt{len}(\bm{a})$, $\rho$ the proportion of $\bm{a}$ for planning, the world model $\Omega$, the planning sequence $\bm{a'}=[]$

\FOR{$t\in\{1, 2, ..., (1-\rho)\cdot N\}$}
\STATE {
$s_{t+1}, r_{t}=\texttt{env}(a_{t}), \bm{a}'.\texttt{append}(a_{t})$\label{alg:policy_planning_env}}
\ENDFOR
\FOR{$t\in \{(1-\rho)N, \dots, (1+\rho)N\}$}
\STATE
$a_{t}=\Omega(s_{t})$
\STATE $s_{t+1}, r_{t}, d_{t}=\Omega(s_{t}, a_{t}), \bm{a}'.\texttt{append}(a_{t})$\label{alg:policy_planning_world_model}\;
\STATE \textbf{if} {$d_{t}$} \textbf{then} \textbf{break}
\ENDFOR

\RETURN $\bm{a}'$
\end{algorithmic}
\end{algorithm}
\end{minipage}
\vspace{-10pt}
\end{figure}

\subsection{Task II: Action Proposal}

\textbf{Task Description.} {The action proposal is a novel task for world model, based on \textbf{Observations 3.1 and 4.2}. Basically, we will ask the world model to recommend top-$K$ actions that can potentially complete the task. Specifically, we follow the representation of the state in the next state prediction, with the additional information: i) the examples of actions, and ii) the previous actions. The previous actions can help the LLMs to understand the game progress.}  The code to LLM for the action proposal is displayed in Appendix~\ref{app:prompt_action_proposal}. One key issue for the action proposal is that the action generated by the world model may not be valid for the game at the current state. Therefore, given the predicted action $a'$ and the set of possible actions to be executed at the current state $A'$, we leverage the text-embedding model, i.e., \texttt{text-embedding-3-small}~\citep{openaiada} to query the most similar actions with the cosine similarity, i.e., $a^*=\arg\max\{\texttt{emb}(a', a), \forall a\in A'\}$.

\textbf{Evaluation Protocol.} The action proposal requires the world model to generate the top-$K$ potential actions to complete the tasks, where $K\in\{1, 2, 3, 5, 10\}$. Specifically, given the action sequence $\bm{a}$ generated by the rule-based policies\footnote{{We note that there may be more than one policy to complete the task. However, we cannot enumerate all possible policies and we view this evaluation as the lower bound performance of the model in action proposal.}}, we will let the world model to generate the potential actions with the states along with the path of $\bm{a}$ to complete the task. We say the action proposal is correct if the actions in $\bm{a}$ in the generated actions by the world model. The results of the accuracy are averaged over the steps over the action sequence and 30 runs of each environment. The action sequence $\bm{a}$ generated by the rule-based policy is not the only sequence to complete the task and we cannot enumerate all possible actions which can lead to the completion of the task. We note that the number of available actions in the environments is usually larger than 500, which brings difficulties to the RL methods for training and indicates the necessity to generate potential actions to facilitate the learning.

\textbf{Evaluation Results.} {The action proposal results are displayed in Figure~\ref{fig:action_proposal}. Overall, GPT-4o consistently outperforms GPT-4o-mini across different tasks and different values of $K$. With the increase of the number of steps to complete the tasks, where more analysis of the previous actions is needed to understand the game progress, GPT-4o maintains the better accuracy, while GPT-4o-mini shows a substantial drop of the accuracy. The performance gap between the two models is generally increased when the number of steps to complete the tasks increase. When $K=10$, the accuracy of the action proposal for GPT-4o is very high in most tasks. With approximately 800 possible actions available at each time step, the results demonstrate that GPT-4o effectively identifies and selects relevant actions while filtering out irrelevant ones. This capability shows promising potential for successful task completion. Furthermore, we still observe that both models obtain lower values in the tasks requiring the domain knowledge, i.e., blood and conduct, which is consistent to the observation in~\citep{wang2024language} that LLMs, e.g., GPT-4, is more likely to make errors when scientific knowledge is needed. {We also provide the step accuracy of the action proposal in Appendix~\ref{app:step_accuracy} to illustrate the prediction of the relevant actions along with the steps. We observe that there are some key steps that has extremely low accuracies, which indicates that the critical steps significantly influences the difficulties of the tasks, rather than the number of steps to complete the tasks, which differs from the traditional RL.} {We also observe that both GPT-4o and GPT-4o-mini can generate wrong actions even when the action rules are given, especially for the environments where the domain scientific knowledge is needed, e.g., `space-walk'.}

\begin{figure*}[t]
\centering
\vspace{-20pt}
    \includegraphics[width=0.95\textwidth]{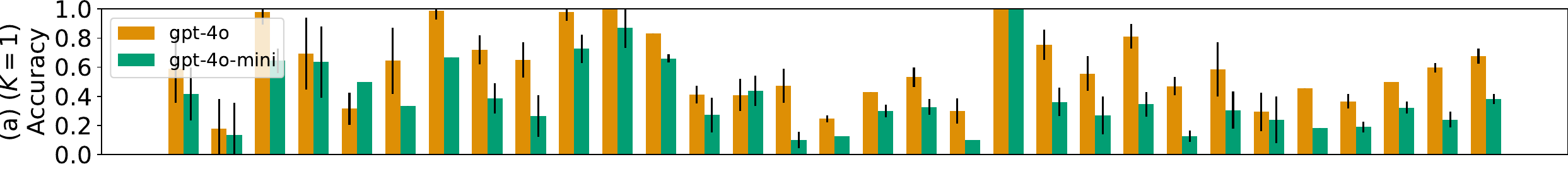}
    \includegraphics[width=0.95\textwidth]{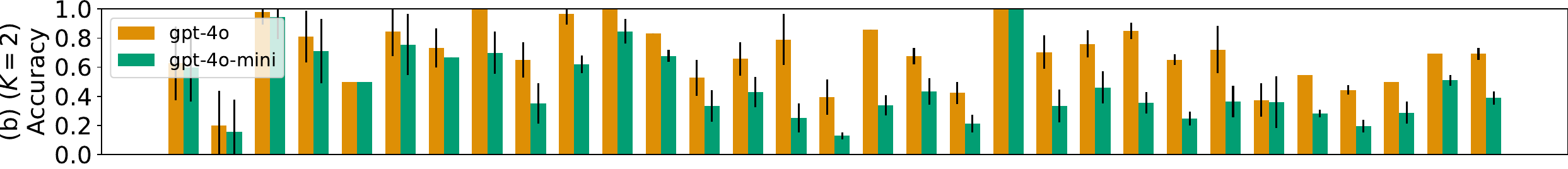}
    \includegraphics[width=0.95\textwidth]{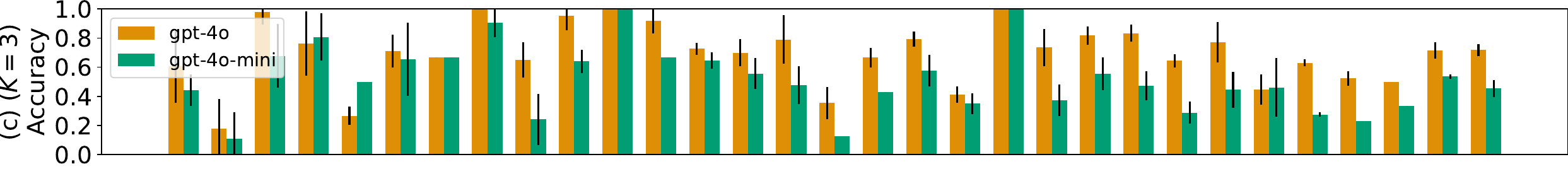}

    \includegraphics[width=0.95\textwidth]{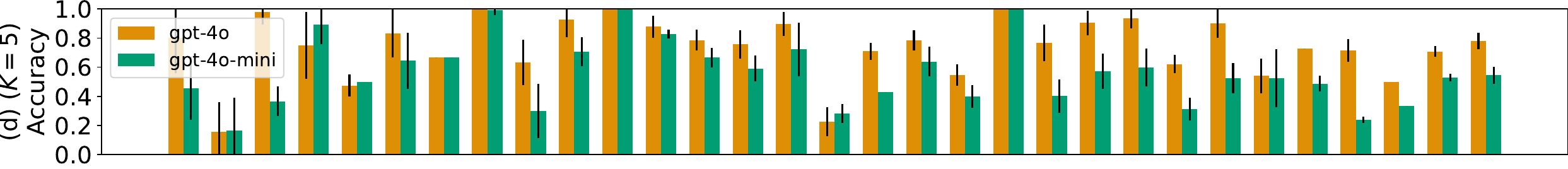}

    \includegraphics[width=0.95\textwidth]{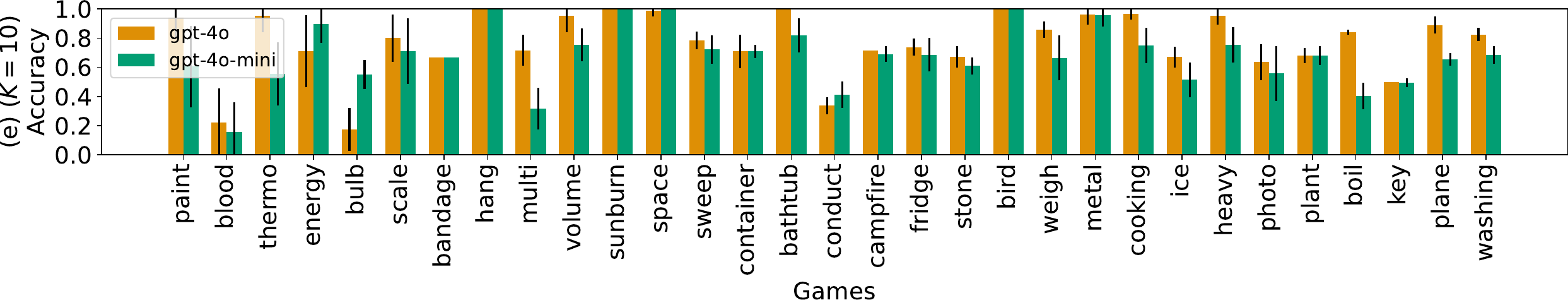}
\vspace{-10pt}
\caption{The accuracy of the world model to generate the potential actions}
\label{fig:action_proposal}
\vspace{-10pt}
\end{figure*}


\begin{tcolorbox}[colback=blue!10, colframe=black, coltitle=black, left=0pt, right=0pt, top=0pt, bottom=0pt, arc=0.5mm, boxrule=0.5pt
]
\textbf{Takeaways}
\begin{itemize}[leftmargin=*, topsep=0pt, partopsep=0pt, parsep=0pt, itemsep=0pt]
\item Action proposal evaluates the world model's ability to generate potential actions that can lead to task completion, going beyond traditional state and reward prediction
\item Performance improves substantially when proposing more candidate actions 
\item Critical steps with extremely low accuracy indicate that specific decision points drive task difficulty more than overall step count, which differs from the traditional decision making
\end{itemize}
\end{tcolorbox}

\subsection{Task III: Policy Planning} 
\label{sec:policy_planning}



\textbf{Task Description.}
The policy planning task is motivated by \textbf{Observations 3.2 and 4.3}, which combines the policy verification and the action proposal (as displayed in Figure~\ref{fig:task_tax}). The process of policy planning is displayed in Algorithm~\ref{alg:policy_planning}. Specifically, we execute the actions in the given $\bm{a}$ on the environment for $(1-\rho) N$ steps (Line~\ref{alg:policy_planning_env} in Algorithm~\ref{alg:policy_planning}) and then plan for $2\rho N$\footnote{The budget $2\rho N$ allows a safety factor of 2 to accommodate potential backtracking or dead-ends in the generated policy. In practice, the loop terminates early when the world model predicts task completion.} steps, i.e., $\{(1-\rho)N, \dots, (1+\rho)N\}$ with the world model (Line~\ref{alg:policy_planning_world_model} in Algorithm~\ref{alg:policy_planning}), where both the action to execute and the state transitions are generated by the world model. The returned action sequence $\bm{a}'$ will be evaluated in the environment to verify the correctness. Only top-1 action is generated in Algorithm~\ref{alg:policy_planning} for illustration. When more actions are generated, we need to enumerate all possible outcomes or leverage search methods, which will be tackled in future work.

\textbf{Evaluation Protocol.} The policy planning is based on the policy verification and the action proposal, as showed in Algorithm~\ref{alg:policy_planning}. Similar to the policy verification, we let $\rho \in \{0.25, 0.5, 0.75, 1.0\}$ to vary the number of steps for the planning. We only consider the case with $K=1$, i.e., the world model only generates the top-1 action with the given states. Finally, we evaluate the planned policy $\bm{a}'$ in the environment to verify the correctness. We note that when $K=1$, no advanced search method is needed, while when $K>1$, we cannot enumerate all possible outcomes for larger steps, e.g., 10. Besides, a critic is also needed to choose among the outcomes for verifying in the environments. Therefore, we only consider the case with $K=1$ and leave the case $K>1$ into future work.

\textbf{Evaluation Results.} {The policy planning results are displayed in Figure~\ref{fig:planning}, where GPT-4o and GPT-4o-mini achieve comparable performance for the tasks with less steps and smaller values of $\rho$, e.g., $0.25$ and GPT-4o generally achieve better results in tasks with more steps. When the value of $\rho$ increases, the performance is generally decreasing. With the coupling of the policy verification and action proposal, we observe more unstabilities of the performances of models over tasks and settings. This indicates the necessity of decoupling the functionalities of the world model for the evaluation. Similar to the model-based RL~\citep{schrittwieser2020mastering}, where introducing the world model may bring the training unstabilities, we need to be carefully to apply the LLM-based world models to the decision making due to the inherent complexity. 
During the experiments, we also observe the format errors of the outputs from both models, which may interrupt the running of the experiments. {The frequency of these failures depends on the environments, where the `hang-painting', `space-walk', and `make-campfire' are the three environments we experiences most of the failures.} Therefore, with the interaction of the different functionalities of the world model, the system is more unstable.}


\begin{figure*}[t]
\centering
\vspace{-20pt}
    \includegraphics[width=.95\textwidth]{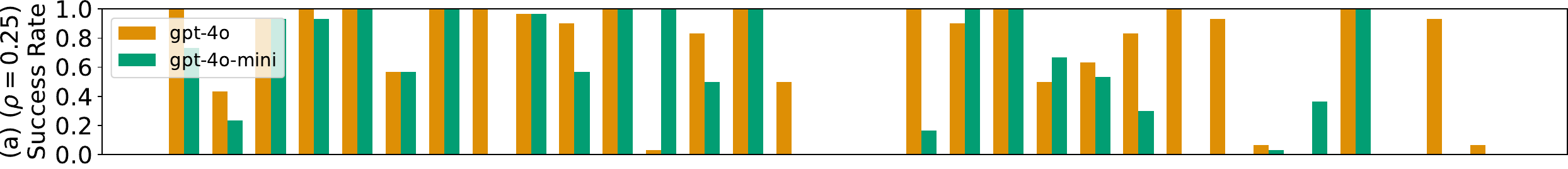}
    \includegraphics[width=.95\textwidth]{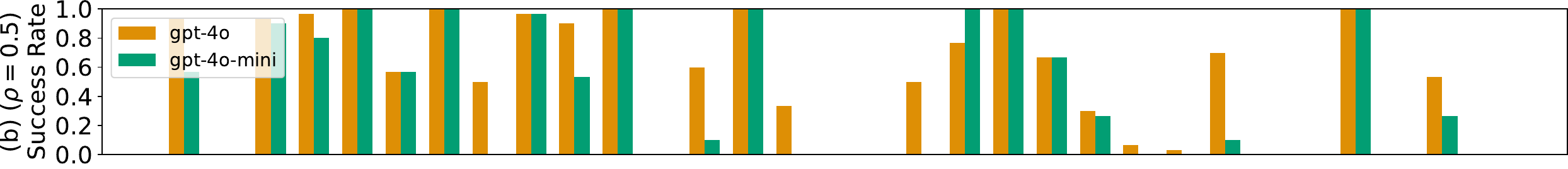}
    \includegraphics[width=.95\textwidth]{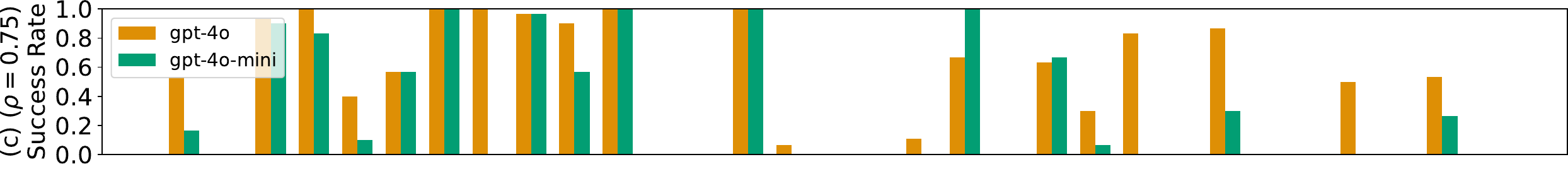}

    \includegraphics[width=.95\textwidth]{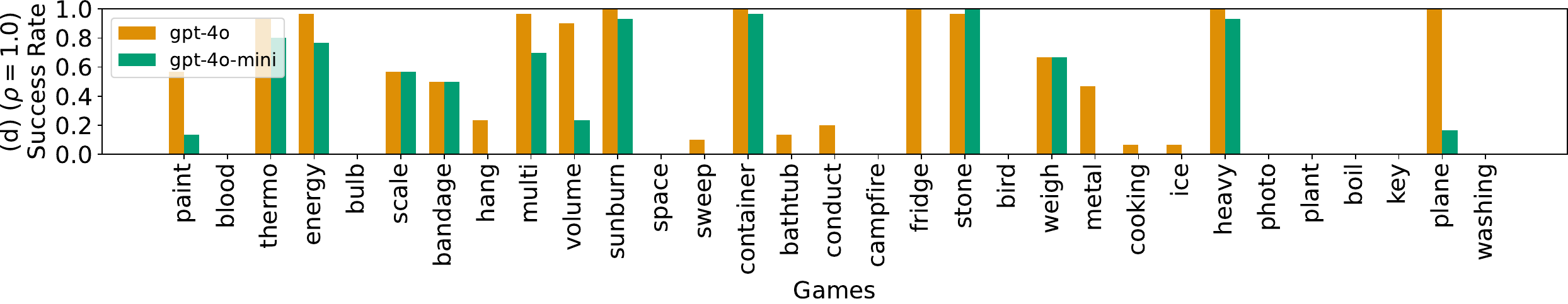}

\vspace{-10pt}
\caption{The success rate of the world model to complete the tasks}
\label{fig:planning}
\vspace{-15pt}
\end{figure*}

\begin{tcolorbox}[colback=blue!10, colframe=black, coltitle=black, left=0pt, right=0pt, top=0pt, bottom=0pt, arc=0.5mm, boxrule=0.5pt
]
\textbf{Takeaways}
\begin{itemize}[leftmargin=*, topsep=0pt, partopsep=0pt, parsep=0pt, itemsep=0pt]
\item Policy planning demonstrates that world models can make decisions solely by combining policy verification and action proposal without requiring separate critic modules
\item Performance becomes more unstable when coupling multiple world model functionalities, partially obscuring performance gaps between strong and weak models
\end{itemize}
\end{tcolorbox}

\vspace{-8pt}
\subsection{Summary}

The three evaluation tasks work together to validate our framework: Policy verification (Task I) confirms that world models can assess multi-step consequences, addressing Observations 4.1 and 4.2. Action proposal (Task II) demonstrates capability to identify promising actions, addressing Observation 3.1. Policy planning (Task III) shows these capabilities can be combined for autonomous decision making without separate actors, addressing Observation 3.2, though with increased instability as suggested by Observation 4.3. Together, these results support our position that LLM-based world models can make decisions solely, while highlighting the importance of decoupled evaluation to understand component capabilities and failure modes. We believe these results validate the reasonability of the suggested evaluation framework and illuminate future research directions for understanding LLM-based world models in decision-making holistically. 

\section{Related Work}
\label{app:related_work}

\textbf{World Models in Decision Making.} World models are actively explored by researchers to further improve the agent's performance and the sample efficiency~\citep{ha2018world,janner2019trust,hafner2019dream,schrittwieser2020mastering}. 
Dreamer~\citep{hafner2019dream,hafner2025training,hafner2025mastering} is a practical model-based reinforcement learning algorithm that introduces the belief over states as a part of the input to
the model-free DRL algorithm used. Trajectory Transformer~\citep{janner2021offline} trains the transformer to prediction the next state and action as a sequence modeling problem for continuous robot control. MuZero~\citep{schrittwieser2020mastering} is a remarkable success of model-based RL, which learns the world model and conduct the planning in the latent space. MuZero achieves the superior performances over other model-based and model-free RL methods. The world models trained in these methods are problem-specific and cannot be generalized to other problems, which motivates researchers to seek to more generalizable world models, e.g., LLMs as world models. The world model with LLM in~\citep{xiang2023language} is trained to gain the environment knowledge, while maintaining other capabilities of the LLMs. Dynalang~\citep{lin2024learning} proposes the multimodal world model to unify videos and texts for future predictions in decision making. 

\textbf{LLMs as World Simulators.} World simulators are developed to model the dynamics of the world~\citep{bruce2024genie}. LLMs serve as the world simulator due to their generalizability across tasks. Specifically, The LLMs (i.e., GPT-3.5 and GPT-4) is evaluated to predict the state transitions, the game progress and scores with the given object, action, and score rules, where these rules are demonstrated to be crucial to the world model predictions~\citep{wang2024language}. The world models with LLMs in~\citep{xie2024making} need to additionally identify the valid actions. Recently, the Genie series~\citep{bruce2024genie,deepmind2024genie2,deepmind2025genie3} learns controllable world models from unlabeled video data and enables users to interact with generated game-like environments through action inputs, followed by MatrixGame~\citep{zhang2025matrix,he2025matrix} and GameCraft~\citep{li2025hunyuan}. These work demonstrate the (multimodal) LLMs' potential to serve as general world simulators. We move a step further to ask the world model to propose the potential actions to complete the tasks. Both methods focus on next state prediction, which may be not suitable for the evaluation of world models for decision making.


\textbf{LLM-based World Models for Decision Making.} The concept of world model also be explored in the deliberation reasoning of LLMs. Specifically, Reasoning via Planning (RAP)~\citep{hao2023reasoning} leverage the planning methods (e.g., Monte Carlo Tree Search (MCTS)) with the world model with LLMs for plan generation and math reasoning, where LLMs need to predict the next state and the reward to guide the search. Tree of Thought (ToT)~\citep{yao2023tree} implicitly leverage the LLMs as the world model to predict the next state and the reward for the search over different thoughts. Reason for future, act for now (RAFA)~\citep{liu2023reason} combine the planning and reflection with the world model for complex reasoning tasks. Recent work considers LLM-based world models for web agents~\citep{chae2025web,gu2024your}, game agents~\citep{hafner2025training}, and even autonomous driving~\citep{liao2025diffusiondrive}. However, these methods do not focus on the evaluation of world models, and several interdependent modules are coupled for completing the task.



\section{Discussions}

\textbf{Are (LLM-based) World Models Necessary?} Despite the success of MuZero~\citep{schrittwieser2020mastering} and Dreamer~\citep{hafner2025mastering,hafner2025training}, world models remain relatively unpopular in the research community. The main barriers are the added complexity of training separate world models and their lack of generalizability. Although LLM-based world models offer improved generalization capabilities, most current LLM agents operate without them.  This raises an important question: \textbf{are (LLM-based) world models necessary?}
We argue that world models will become critical components of LLM agents for several compelling reasons. First, as agents are deployed in real-world scenarios where actions carry significant consequences, e.g., high-stake scenarios, world models enable essential counterfactual reasoning and outcome prediction prior to execution. This capability allows agents to simulate potential consequences without incurring the costs and risks of trial-and-error in physical environments. Second, world models enhance interpretability by explicitly modeling future states and outcomes, thereby making agent behavior more auditable and transparent to human stakeholders. Third, world models facilitate sample-efficient learning by enabling agents to plan and refine strategies through internal simulation rather than requiring extensive real-world interaction.
Most critically, world models will play an essential role in safety assurance and governance, serving as a foundational safeguard in the development of safe artificial general intelligence (AGI). As we advance toward increasingly autonomous and complicated AI systems, the capacity to accurately model and reason about environmental dynamics and action consequences becomes not merely with better performance, but also necessary for responsible AI deployment in the real world.

\textbf{Limitations.} There are several limitations of the current framework. i) Our framework relies on rule-based policies as ground truth, which represents only one valid solution path per task. Real-world decision making often admits multiple valid strategies. This single-path assumption may penalize world models that propose alternative but potentially successful approaches, leading to underestimation of their true capabilities. ii) For policy planning, we only evaluate $K=1$ due to computational constraints. This limitation prevents us from assessing the world model to maintain consistency across multiple predicted trajectories, which is important for tree search methods. {Extending to K>1 with tree search is important future work to better demonstrate full planning capabilities.} To facilitate the understanding, we provide a comprehensive discussion about the role of $K$ in Appendix~\ref{sec:role_of_K}. iii) While we argue that decoupling evaluation of world model functionalities is critical, this approach may miss important interactions between components. The performance degradation in policy planning suggests that integrated evaluation still provides valuable insights. {iv) Due to the limited budget, we only conduct the experiments on two widely used models. Our claims should be interpreted as applying to the tested models, and validation across diverse model families (including open-source models like Qwen-72B, Llama-70B, and closed models like Claude and Gemini) remains important future work.}

\textbf{Conclusion.} In this work, we argue that LLM-based world models can make decisions solely through the combination of policy verification and action proposal capabilities, but require rigorous, decoupled evaluation to understand their true strengths and limitations. Our proposed bottom-up evaluation framework, i.e., policy verification, action proposal, and policy planning tasks, enables systematic assessment across 31 diverse environments spanning daily and scientific domains. Key findings reveal that: i) GPT-4o significantly outperforms GPT-4o-mini, particularly on domain knowledge-intensive tasks, ii) performance depends predominantly on critical steps rather than total step count, and iii) integrating multiple world model functionalities introduces instability that obscures capability differences between models. While our framework has limitations—including reliance on single-path ground truth policies and restricted evaluation of multi-action planning—it establishes a foundation for systematic world model assessment that shifts focus from general simulation accuracy to decision-oriented performance. As LLM-based agents advance toward deployment in high-stakes real-world scenarios, our evaluation framework becomes essential for ensuring their safe and effective operation in complex decision-making tasks.

\subsubsection*{Acknowledgments}
This work was partially supported by the Singapore Ministry of Education (MOE) Academic Research Fund (AcRF) Tier 1 grant (Proposal ID: 23-SIS-SMU-037).
The work described in this paper was also partially supported by a grant from the Innovation and Technology Commission of the Hong Kong Special Administrative Region, China (Project No. GHP/391/22).

\bibliography{world}
\bibliographystyle{tmlr}

\appendix
\section{Frequently Asked Questions (FAQs)}

\subsection{Why World Models and Why World Models with LLMs?}
i) Generalization to novel tasks: World models have demonstrated impressive transfer learning abilities~\citep{byravan2020imagined}, allowing agents to adapt to previously unseen scenarios by leveraging their learned understanding of world dynamics. This generalization capacity is particularly valuable in robotics and control applications where agents must handle diverse situations~\citep{robey2021model,young2023benefits}.
ii) Efficient planning: The predictive capabilities of world models enable the sophisticated planning algorithms without any online interactions~\citep{sekar2020planning,hamrick2021role,schrittwieser2020mastering}. By simulating possible futures, agents can evaluate different action sequences and select optimal strategies without requiring actual interaction with the environment. This ``imagination'' or ``mental simulation'' capability dramatically improves sample efficiency and safety in decision-making.
iii) Offline learning: World models have proven especially valuable in offline reinforcement learning settings~\citep{schrittwieser2021online,yu2020mopo,yu2021combo}, where agents must learn from pre-collected datasets without direct environment interaction. The ability to learn accurate dynamics models from historical data has opened new possibilities for training agents in scenarios where online interaction is impractical or costly.

\subsection{More Justifications of the Proposed Tasks}
{In this section, we will provide a mode detailed justification of the three tasks proposed in this work.}

\paragraph{Policy Verification.} {Policy verification can be viewed as a generalization of the next state prediction. Instead of focusing on the accuracy of the one-step prediction about the next states and the reward/terminal prediction, which is considered in most previous works, policy verification may accumulate the multi-step predictions and judging whether the given policy can complete the task or not. This task is more relevant to the world model for decision making, as if the world model can verify any given policy correctly, with the enough number of sampling of the policy, i.e., action sequences, we can complete the task in the end.}

\paragraph{Action Proposal.} {As observing in the ToT~\citep{yao2023tree}, generating useful thoughts is critical which can significantly improve the performance. However, with multiple thoughts generated, we have to select one to executed. We can test these thoughts in the environments, however, this is not always doable. Therefore, building a world model is the straightforward way to do this. On the other hand, action proposal is necessary for the world model as the game engine to guide the fresh players to complete the game. With increasing the number of recommend actions, the difficulty of the action proposal is decreased.  However, this task is not considered in the previous work for the world model. We believe that this task should be an important task to evaluate the world model for decision making.}

\paragraph{Policy Planning.} {Policy planning is a combination of policy verification and action proposal. Conceptually, if the world model performs well on both tasks, we can obtain the policy with world model only and no actor is needed. This can help us to understand the world model through decoupling the world model with any other modules. Besides, this planning task is consistent with the System 2 thinking, i.e., with more time for the planning, the world model may find better policies. }

\subsection{The Objectives of This Paper}

{The primary objective of this paper is proposing the new evaluation tasks for the evaluation of the world models with LLMs for decision making. }
\begin{itemize}
    \item {Evaluating world model for decision making is difficult, given that the decision making tasks usually involves multiple steps of the predictions. Therefore, the one-step prediction tasks considered in most previous works is not suitable.}
    \item {Instead of treating the world models as world simulators or supporting modules for the actors, we identify that the world model can solve the tasks solely with the combination of the policy verification and action proposal. Therefore, the world model should be researched with the same importance. }
    \item {World models have traditionally been evaluated through a top-down approach, where complex systems are constructed to complete tasks, constraining analysis to high-level observations. By examining fundamental capabilities like policy verification and action proposal, we propose a bottom-up evaluation that enables more systematic and granular assessment of world models.}
\end{itemize}

\subsection{Selection of Backbone LLMs}

\begin{itemize}
    \item {We need to evaluate the three novel tasks over 31 environments, and each with 30 runs, we roughly use 3000 dollars for all the experiments for GPT-4o and GPT-4o-mini. Due to the limited budget, we cannot afford to test on Claude and Gemini.}
    \item {For the open-sourced models, we test on some open-sourced models, e.g., Qwen 7B, and find that current open-sourced models still cannot generate the responses with correct formats, i.e., JSON. This brings difficulties for the evaluation.}
\end{itemize}

\subsection{The Role of $K$}
\label{sec:role_of_K}

The choice of $K$ in policy planning, i.e., the number of candidate actions proposed at each step, deserves careful discussion, as it governs the trade-off between world model isolation and practical planning performance, as well as the practical efficiency of applying world models for planning.

\paragraph{$K=1$ as a Principled Baseline.}
Our main evaluation is with fixed $K=1$, which represents the most straightforward test of the world model: at each step, the model must commit to a single action with no opportunity for correction. This is a deliberate methodological choice rather than a computational limitation. When $K>1$, the planner must select among candidate sequences using a search strategy, e.g., beam search, majority voting, or MCTS, and success becomes a joint function of world model quality \emph{and} search algorithm design. This reintroduces precisely the coupling problem identified in Observation~4.3, which our decoupled evaluation framework is designed to avoid. $K=1$ therefore provides the cleanest signal of the world model's intrinsic planning capability, and establishes a meaningful lower bound.

\paragraph{What $K>1$ Would Look Like.}
The action proposal results in Section~5.2 provide an indirect but informative window into $K>1$ planning behavior, since action proposal is essentially policy planning evaluated one step at a time. Two observations are relevant. First, at $K=10$, GPT-4o achieves near-perfect per-step coverage on the majority of environments, meaning the correct action is almost always present in the top-10 candidates. The $K=1$ planning bottleneck is therefore greedy commitment at critical steps, not a lack of world model knowledge. A simple $K>1$ selection strategy such as majority voting would be expected to substantially improve planning success rates, particularly on environments where a single bottleneck step has low $K=1$ accuracy but high $K=10$ coverage (e.g., \textit{make-campfire}, \textit{space-walk}). Second, for environments with long sequential dependencies (e.g., \textit{wash-clothes}), increasing $K$ alone is insufficient: an error early in the sequence propagates regardless of how many candidates are generated later. In such cases, $K>1$ planning would need to be paired with policy verification as a subroutine, i.e., generating $K$ candidate continuations and verifying each before executing, which is a natural combination of our framework.

\paragraph{The Instability Concern.}
Our experiment results in Subsection~\ref{sec:policy_planning} show that combining world model functionalities already increases performance instability at $K=1$. $K>1$ planning would amplify this effect: each additional branch multiplies the surface area for compounding prediction errors and format failures. This suggests that the evaluation of $K>1$ planning should proceed carefully, with explicit ablations over search strategy and $K$ separately, rather than treating $K>1$ as a straightforward extension. We leave this systematic study to future work, but emphasize that our framework directly supports it, i.e., policy verification and action proposal are already evaluated as independent components, and their combination under $K>1$ search is a well-defined next step.

\paragraph{Summary.}
$K=1$ is the starting point for understanding world model capability in isolation and $K>1$ is the next step for understanding world model capability in practice. The gap between them is bridged by search strategy design, which we treat as an orthogonal research question. Our results suggest that this gap is largely attributed to bottleneck steps rather than general knowledge deficiency, and that policy verification is the natural mechanism for exploiting $K>1$ candidates effectively.

\subsection{{Comparison with Decision Transformer and Trajectory Transformer}}

{Decision transformer (DT)~\citep{chen2021decision} trains the transformer to predict the action conditional on the experiences and the target reward or the goal. Language DT (LDT)~\citep{gontier2023language} extends DT to consider the text-based games and include the state prediction in the training as an auxiliary tasks. However, during the inference, i.e., decision making, the model still generates the action directly, which is not based on the world model because the state prediction is only used for training and not for acting.}

{Trajectory Transformer (TT)~\citep{janner2021offline} also consider the decision making problem as a sequence modeling problem, where the transformer is trained to predict the state, the action and the reward. Compared with DT, TT is more related to the world model, where the state and reward prediction is used to generating actions and the search methods, e.g., beam search, is used. However, only the continuous robot control is considered in TT~\citep{janner2021offline} and the trained TT model is domain and problem specific, which cannot generalize to other problems. }

{Recently, LLMs provide a promising way to build the general world model and the world model with LLMs emerge as a novel research field. However, most of these work focus on the single-step prediction and a comprehensive evaluate is needed. This work is inspired by TT and extends the insights from TT to the world model with LLMs for text-based games. Specifically, we consider the policy verification, the action proposal and the policy planning tasks, where the TT combines these tasks to generate the actions and only investigate the performance for the decision. Instead of only considering the performance of the decision makings, our three tasks provide a bottom-up analysis of the world model for decision making. }

\subsection{Broader Impact Statement}
{This work contributes an evaluation methodology for LLM-based world models towards AI safety and responsible development. Rigorous evaluation frameworks help prevent insecure deployment of vulnerable systems in high-stakes scenarios. However, we acknowledge potential dual-use concerns: improved world models could enhance autonomous agent capabilities, which could be misused for harmful automation. We emphasize that our work is evaluation-oriented and that substantial research on safety, robustness, and value alignment remains necessary before deploying world-model-based agents in real-world applications.}

\subsection{Code and Dataset Availability}

The code and datasets can be access at: \url{https://github.com/joannacyang/WorldModel_TMLR}.

\clearpage
\section{Environments}

\subsection{Introduction of Tasks}
\label{app:env_task}
\begin{table}[ht]
\centering
\caption{Environments~\citep{wang2024language}}
\label{tab:envs}
\begin{tabular}{m{4cm}|m{9cm}}
\toprule 
Environments & Task Description\\
\midrule
mix-paint (paint)	&	Your task is to use chemistry to create black paint.\\
blood-type (blood)	&	Your task is to give a correct type of blood to the patient.\\
thermometer (thermo)	&	Your task is to figure out the temperature of the water in the pot.\\
clean-energy (energy)	&	Your task is to change all fossil-fuel power stations to use renewable energy while keeping the same capacity.\\
lit-lightbulb (bulb)	&	Your task is to lit the light bulb.\\
scale-weigh (scale)	&	Your task is to figure out the weight of the apple.\\
use-bandage (bandage)	&	Your task is to put bandages on any cuts.\\
hang-painting (hang)	&	Your task is to hang the picture of a girl (ID: 11) on the back wall (ID: 5).\\
multimeter (multi)	&	Your task is to figure out the resistance of the resistor 0.\\
volume (volume)	&	Your task is to figure out the volume of the green box.\\
sunburn (sunburn)	&	It is a summer noon. The sky is clear. Your task is to take a ball from the beach and put it in the box in the house. Protect yourself from sunburn!\\
space-walk (space)	&	Your task is to conduct a space walk.\\
sweep-floor (sweep)	&	Your task is to clean the garbage on the ground to the garbage can.\\
volume-container (container)	&	Your task is to figure out the volume of the glass.\\
bath-tub-water-temperature (bathtub)	&	Your task is to make the temperature of the water in the bath tub to 35 - 40 Celsius degree by adding water from the taps. When you are done, take the action "bath".\\
conductivity (conduct)	&	Your task is to figure out if the fork is conductive or not. If the fork is conductive, put it in the red box. Otherwise, put it in the black box.\\
make-campfire (campfire)	&	Your task is to make a fire in the fire pit.\\
refrigerate-food (fridge)	&	Your task is to prevent the foods from spoiling.\\
volume-stone (stone)	&	Your task is to figure out the volume of the stone.\\
bird-life-cycle (bird)	&	Your task is to hatch the egg and raise the baby bird.\\
balance-scale-weigh (weigh)	&	Your task is to figure out the weight of the cube. Use the answer action to give your answer.\\
metal-detector (metal)	&	Your task is to find the buried metal case on the beach. You win the game by putting the metal case in your inventory.\\
cooking (cooking)	&	Your task is to prepare a meal following the instructions of the cook book.\\
make-ice-cubes (ice)	&	Your task is to make ice cubes.\\
balance-scale-heaviest (heavy)	&	Your task is to put all heaviest cubes into the box.\\
take-photo (photo)	&	Your task is to take a nice picture of orange (ID: 4), using a camera with shutter speed of 1/2, aperture of 16, and iso of 1600.\\
plant-tree (plant)	&	Your task is to plant the tree and water it.\\
boil-water (boil)	&	Your task is to boil water.\\
forge-key (key)	&	Your task is to forge a key to open the door.\\
inclined-plane (plane)	&	Here are two inclined planes with the same angle. Your task is figure out which of the two inclined planes has the most friction. Focus on the inclined plane with the most friction after your experiment.\\
wash-clothes (washing)	&	Your task is to wash the dirty clothes and dry them.\\
\bottomrule
\end{tabular}

\end{table}
{There are \textbf{32} environment in~\citep{wang2023bytesized} and dish-washing is selected as the example in the prompt, which is excluded for fair evaluation. Specifically, the environments can be categorized into two domains:
\begin{itemize}
    \item \textbf{Daily-life tasks}, including use-bandage, hang-painting, sunburn, sweep-floor, bath-tub-water-temperature, make-campfire, refrigerate-food, cooking, take-photo, plant-tree, boil-water, and wash-clothes. For these tasks, the world model need to have the common knowledge about the procedure of completing these tasks, e.g.. first collecting the dirty clothes, then put them into the washing machine, then use the detergent and turn on the washing machine for the wash-clothes task.
    \item \textbf{Scientific tasks}, including mix-paint, blood-type, thermometer, clean-energy, lit-lightbulb, scale-weigh, multimeter, volume, space-walk, volume-container, conductivity, volume-stone, bird-life-cycle, balance-scale-weigh, metal-detector, make-ice-cubes, forge-key and inclined-plan. These tasks requires the scientific knowledge to complete the tasks, e.g., the world model need to know that the friction may decrease the speed of a block sliding down of the plane for the inclined-plane. Then, the world model need to generate build a micro-simulation to compare the frictions of the two planes.
\end{itemize}

\subsection{Code for Demo Actions Generation}
\label{app:rule_based_policies}
Only one playthrough of the game is provided in~\citep{wang2024language}, which is not enough due to the randomness in the environments. Therefore, we curate the rule-based policy for each environment. Note that due to the randomness of the environments, e.g., the target color in the mix-paint task, the generated action sequences are different for different instantances of the same environment.

\begin{lstlisting}[breaklines = true, caption = {mix-paint}]
    def get_demo_actions(self):
        target_color = self.useful_info[0][0]
        paint_names = self.useful_info[1]

        color_dict = {
            # "red": (1, 0, 0),
            "orange": ["red", "yellow"],
            # "yellow": (0, 1, 0),
            "green": ["yellow", "blue"],
            # "blue": (0, 0, 1),
            "purple": ["red", "blue"],
            "black": ["red", "yellow", "blue"],
        }
        paint_names_to_idx = {}
        for paint_idx, paint_name in enumerate(paint_names):
            paint_names_to_idx[paint_name] = paint_idx
        color_mix_plan = color_dict[target_color]

        demo_actions = []
        to_idx = -1
        for color_idx, color_mix in enumerate(color_mix_plan):
            if color_idx == 0:
                to_idx = paint_names_to_idx[color_mix]
                continue
            demo_actions.append(
                "pour {} paint (ID: {}) in cup {} (ID: {})".format(
                    color_mix,
                    2 * (paint_names_to_idx[color_mix] + 1) + 1,
                    to_idx,
                    2 * (to_idx + 1),
                )
            )
        demo_actions.append("mix cup {} (ID: {})".format(to_idx, 2 * (to_idx + 1)))

        return demo_actions
\end{lstlisting}
\begin{lstlisting}[breaklines = true, caption = {blood-type}]
    def get_demo_actions(self):
        useful_info = self.useful_info[1]

        return [
            "give Type {} {} blood (ID: 3) to patient (ID: 2)".format(
                useful_info[0], useful_info[1]
            ),
            "take Type {} {} blood (ID: 3)".format(useful_info[0], useful_info[1]),
            "give Type {} {} blood (ID: 3) to patient (ID: 2)".format(
                useful_info[0], useful_info[1]
            ),
        ]
\end{lstlisting}
\begin{lstlisting}[breaklines = true, caption = {thermometer}]
    def get_demo_actions(self):
        demo_actions = [
            "take thermometer (ID: 4)",
            "use thermometer (ID: 4) on water (ID: 3)",
            "answer {} Celsius degree".format(self.water_temperature),
        ]
        return demo_actions
\end{lstlisting}
\begin{lstlisting}[breaklines = true, caption = {clean-energy}]
    def get_demo_actions(self):
        demo_actions = []
        change_station = {
            "sun": "solar farm",
            "water": "hydroelectric power station",
            "wind": "wind farm",
        }
        for region in self.regions:
            demo_actions.append(
                "change {} to {}".format(
                    region.name, change_station[region.properties["resource"]]
                )
            )

        return demo_actions
\end{lstlisting}
\begin{lstlisting}[breaklines = true, caption = {lit-lightbulb}]
    def get_demo_actions(self):
        return [
            "connect light bulb (ID: 2) terminal1 to red wire (ID: 3) terminal1",
            "connect red wire (ID: 3) terminal2 to battery (ID: 6) anode",
            "connect battery (ID: 6) cathode to black wire (ID: 4) terminal1",
            "connect black wire (ID: 4) terminal2 to light bulb (ID: 2) terminal2",
        ]
\end{lstlisting}
\begin{lstlisting}[breaklines = true, caption = {scale-weigh}]
    def get_demo_actions(self):
        demo_actions = [
            "take {}".format(self.useful_info[0].name),
            "put {} on {}".format(self.useful_info[0].name, self.useful_info[1].name),
            "look",
            "answer {}g".format(self.target_weight),
        ]

        return demo_actions
\end{lstlisting}
\begin{lstlisting}[breaklines = true, caption = {use-bandage}]
    def get_demo_actions(self):
        demo_actions = [
            "open bandage box (ID: 8)",
            "look",
            "take bandage (ID: 9)",
        ]
        return demo_actions + [
            "put bandage (ID: 9) on {} (ID: 3)".format(self.useful_info[0])
        ]
\end{lstlisting}
\begin{lstlisting}[breaklines = true, caption = {hang-painting}]
    def get_demo_actions(self):
        demo_actions = [
            "take nail (ID: 7)",
            "take hammer (ID: 6)",
            "hammer nail (ID: 7) on {} with hammer (ID: 6)".format(
                self.target_wall.name
            ),
            "take {}".format(self.target_picture.name),
            "hang {} on nail (ID: 7)".format(self.target_picture.name),
        ]

        return demo_actions
\end{lstlisting}
\begin{lstlisting}[breaklines = true, caption = {multimeter}]
    def get_demo_actions(self):
        demo_actions = [
            "set multimeter (ID: 2) to resistance mode",
            "connect multimeter (ID: 2) terminal1 to resistor {} (ID: 3) terminal1".format(
                self.target_resistor_id
            ),
            "connect multimeter (ID: 2) terminal2 to resistor {} (ID: 3) terminal2".format(
                self.target_resistor_id
            ),
            "look",
            "answer {} ohm".format(self.target_resistance),
        ]

        return demo_actions
\end{lstlisting}
\begin{lstlisting}[breaklines = true, caption = {volume}]
    def get_demo_actions(self):
        demo_actions = [
            "take {}".format(self.useful_info[1].name),
            "measure the length of the {} with the {}".format(
                self.useful_info[0].name, self.useful_info[1].name
            ),
            "measure the width of the {} with the {}".format(
                self.useful_info[0].name, self.useful_info[1].name
            ),
            "measure the height of the {} with the {}".format(
                self.useful_info[0].name, self.useful_info[1].name
            ),
            "answer {} cubic cm".format(self.target_box_volume),
        ]

        return demo_actions
\end{lstlisting}
\begin{lstlisting}[breaklines = true, caption = {sunburn}]
    def get_demo_actions(self):
        return [
            "use sunscreen (ID: 4)",
            "move to beach (ID: 3)",
            "look",
            "take ball (ID: 8)",
            "move to house (ID: 2)",
            "put ball (ID: 8) in box (ID: 5)",
        ]
\end{lstlisting}
\begin{lstlisting}[breaklines = true, caption = {space-walk}]
    def get_demo_actions(self):
        return [
            "put on space suit (ID: 7)",
            "open inner door (ID: 5)",
            "move to airlock (ID: 3)",
            "look",
            "close inner door (ID: 5)",
            "open outer door (ID: 6)",
            "move to outer space (ID: 4)",
        ]
\end{lstlisting}
\begin{lstlisting}[breaklines = true, caption = {sweep-floor}]
    def get_demo_actions(self):
        sweep_actions = []
        for garbage in self.useful_info:
            sweep_actions.append(
                "sweep {} to dustpan (ID: 3) with broom (ID: 2)".format(garbage.name)
            )

        demo_actions = (
            ["take broom (ID: 2)", "take dustpan (ID: 3)"]
            + sweep_actions
            + [
                "open garbage can (ID: 4)",
                "dump dustpan (ID: 3) to garbage can (ID: 4)",
            ]
        )

        return demo_actions
\end{lstlisting}
\begin{lstlisting}[breaklines = true, caption = {volume-container}]
    def get_demo_actions(self):
        demo_actions = [
            "take {}".format(self.useful_info[0].name),
            "put {} in sink (ID: 2)".format(self.useful_info[0].name),
            "turn on sink (ID: 2)",
            "turn off sink (ID: 2)",
            "take {}".format(self.useful_info[0].name),
            "pour water in {} into {}".format(
                self.useful_info[0].name, self.useful_info[1].name
            ),
            "look",
            "answer {} mL".format(self.target_water_container_volume),
        ]

        return demo_actions
\end{lstlisting}
\begin{lstlisting}[breaklines = true, caption = {bath-tub-water-temperature}]
    def get_demo_actions(self):

        water_temp = self.useful_info[0]

        cooling = [
            "turn on cold tap (ID: 5)",
            "turn off cold tap (ID: 5)",
            "use thermometer (ID: 6) on water (ID: 3)",
        ]
        hotting = [
            "turn on hot tap (ID: 4)",
            "turn off hot tap (ID: 4)",
            "use thermometer (ID: 6) on water (ID: 3)",
        ]

        if water_temp > 40:
            cooling_times = (water_temp - 35) // 5
            water_actions = cooling * cooling_times

        elif water_temp < 35:
            hotting_times = (40 - water_temp) // 5
            water_actions = hotting * hotting_times
        else:
            water_actions = []

        demo_actions = (
            [
                "take thermometer (ID: 6)",
                "use thermometer (ID: 6) on water (ID: 3)",
            ]
            + water_actions
            + ["bath"]
        )

        return demo_actions
\end{lstlisting}
\begin{lstlisting}[breaklines = true, caption = {conductivity}]
    def get_demo_actions(self):
        demo_actions = [
            "connect light bulb (ID: 2) terminal1 to red wire (ID: 3) terminal1",
            "connect red wire (ID: 3) terminal2 to battery (ID: 6) anode",
            "connect battery (ID: 6) cathode to black wire (ID: 4) terminal1",
            "connect black wire (ID: 4) terminal2 to fork (ID: 7) terminal1",
            "connect fork (ID: 7) terminal2 to blue wire (ID: 5) terminal1",
            "connect blue wire (ID: 5) terminal2 to light bulb (ID: 2) terminal2",
            "look",
            "take fork (ID: 7)",
        ]
        if self.useful_info[0]:
            return demo_actions + ["put fork (ID: 7) in red box (ID: 8)"]

        else:
            return demo_actions + ["put fork (ID: 7) in black box (ID: 9)"]
\end{lstlisting}
\begin{lstlisting}[breaklines = true, caption = {make-campfire}]
    def get_demo_actions(self):
        return [
            "take axe (ID: 4)",
            "use axe (ID: 4) on tree (ID: 5)",
            "look",
            "use axe (ID: 4) on chopped down tree (ID: 5)",
            "look",
            "take firewood (ID: 5)",
            "put firewood (ID: 5) in fire pit (ID: 2)",
            "take match (ID: 3)",
            "use match (ID: 3) on firewood (ID: 5)",
        ]
\end{lstlisting}
\begin{lstlisting}[breaklines = true, caption = {refrigerate-food}]
    def get_demo_actions(self):
        take_objects = []

        put_objects = []
        for food in self.useful_info:
            take_objects.append("take {}".format(food.name))
            put_objects.append("put {} in fridge (ID: 2)".format(food.name))

        demo_actions = (
            take_objects
            + ["open fridge (ID: 2)"]
            + put_objects
            + [
                "close fridge (ID: 2)",
                "look",
                "look",
                "look",
            ]
        )

        return demo_actions
\end{lstlisting}
\begin{lstlisting}[breaklines = true, caption = {volume-stone}]
    def get_demo_actions(self):
        demo_actions = [
            "take measuring cup (ID: 4)",
            "put measuring cup (ID: 4) in sink (ID: 2)",
            "turn on sink (ID: 2)",
            "turn off sink (ID: 2)",
            "take measuring cup (ID: 4)",
            "examine measuring cup (ID: 4)",
            "take stone (ID: 3)",
            "put stone (ID: 3) in measuring cup (ID: 4)",
            "examine measuring cup (ID: 4)",
            "answer {}".format(self.answer_volume),
        ]

        return demo_actions
\end{lstlisting}
\begin{lstlisting}[breaklines = true, caption = {bird-life-cycle}]
    def get_demo_actions(self):
        return [
            "sit on egg",
            "sit on egg",
            "sit on egg",
            "sit on egg",
            "sit on egg",
            "feed young bird",
            "feed young bird",
            "feed young bird",
            "feed young bird",
            "feed young bird",
        ]
\end{lstlisting}
\begin{lstlisting}[breaklines = true, caption = {balance-scale-weigh}]
    def get_demo_actions(self):
        weight_list = [1, 1, 2, 5, 10]
        weight_list_index = [False] * 5

        def _find_combination():
            remaining = self.cube_weight

            # using 10
            if remaining >= 10:
                weight_list_index[-1] = True
                remaining -= 10

            # using 5
            if remaining >= 5:
                weight_list_index[-2] = True
                remaining -= 5

            # using 2
            if remaining >= 2:
                weight_list_index[-3] = True
                remaining -= 2

            if remaining > 0:
                if remaining == 1:
                    weight_list_index[0] = True
                if remaining == 2:
                    weight_list_index[0] = True
                    weight_list_index[1] = True

        _find_combination()

        weight_actions = []
        for idx, weight_list_idx in enumerate(weight_list_index):
            if weight_list_idx:
                weight_actions += [
                    "take {}".format(self.useful_info[idx].name),
                    "put {} in right side of the balance scale (ID: 4)".format(
                        self.useful_info[idx].name
                    ),
                    "look",
                ]

        demo_actions = (
            [
                "take cube (ID: 10)",
                "put cube (ID: 10) in left side of the balance scale (ID: 3)",
            ]
            + weight_actions
            + ["answer {}g".format(self.cube_weight)]
        )

        return demo_actions
\end{lstlisting}
\begin{lstlisting}[breaklines = true, caption = {metal-detector}]
    def get_demo_actions(self):
        agent_init_position = self.useful_info[0]
        targe_position = self.useful_info[1]

        direction = (
            targe_position[0] - agent_init_position[0],
            targe_position[1] - agent_init_position[1],
        )
        h_dir_list = (
            ["south"] * direction[0]
            if direction[0] > 0
            else ["north"] * (-direction[0])
        )
        v_dir_list = (
            ["east"] * direction[1] if direction[1] > 0 else ["west"] * (-direction[1])
        )

        dir_list = h_dir_list + v_dir_list
        self.random.shuffle(dir_list)

        detect_actions = [
            "detect with metal detector (ID: 15)",
        ]
        for dir_step in dir_list:
            detect_actions.append("move {}".format(dir_step))
            detect_actions.append("detect with metal detector (ID: 15)")

        demo_actions = (
            ["take metal detector (ID: 15)", "take shovel (ID: 16)"]
            + detect_actions
            + ["look", "dig with shovel (ID: 16)", "look", "take metal case (ID: 11)"]
        )

        return demo_actions
\end{lstlisting}
\begin{lstlisting}[breaklines = true, caption = {cooking}]
    def get_demo_actions(self):
        cooking_actions = [
        ]

        for cooking_item in self.receipt:
            operations = self.receipt[cooking_item]
            cooking_actions += ["take {}".format(cooking_item.name)]
            for operation in operations:
                if operation in ["slice", "dice", "chop"]:
                    cooking_actions += [
                        "{} {} with {}".format(
                            operation, cooking_item.name, self.useful_info["knife"].name
                        )
                    ]
                if operation in ["fry"]:
                    cooking_actions += [
                        "cook {} in {}".format(
                            cooking_item.name, self.useful_info["stove"].name
                        )
                    ]
                if operation in ["roast"]:
                    cooking_actions += [
                        "cook {} in {}".format(
                            cooking_item.name, self.useful_info["oven"].name
                        )
                    ]

        demo_actions = (
            [
                "take {}".format(self.useful_info["cook_book"].name),
                "read {}".format(self.useful_info["cook_book"].name),
                "take {}".format(self.useful_info["knife"].name),
            ]
            + cooking_actions
            + [
                "prepare meal",
            ]
        )

        return demo_actions
\end{lstlisting}
\begin{lstlisting}[breaklines = true, caption = {make-ice-cubes}]
    def get_demo_actions(self):
        return [
            "open freezer (ID: 2)",
            "examine freezer (ID: 2)",
            "take ice cube tray (ID: 3)",
            "put ice cube tray (ID: 3) in sink (ID: 4)",
            "turn on sink (ID: 4)",
            "turn off sink (ID: 4)",
            "take ice cube tray (ID: 3)",
            "put ice cube tray (ID: 3) in freezer (ID: 2)",
            "close freezer (ID: 2)",
            "look",
            "look",
            "look",
        ]
\end{lstlisting}
\begin{lstlisting}[breaklines = true, caption = {balance-scale-heaviest}]
    def get_demo_actions(self):

        demo_actions = [
            "take {}".format(self.useful_info[0][0].name),
            "put {} in left side of the balance scale (ID: 3)".format(
                self.useful_info[0][0].name
            ),
            "take {}".format(self.useful_info[1][0].name),
            "put {} in right side of the balance scale (ID: 4)".format(
                self.useful_info[1][0].name
            ),
            "look",
        ]

        if len(self.useful_info) == 2:
            if self.useful_info[0][1] > self.useful_info[1][1]:
                # left is heavier

                demo_actions.append("take {}".format(self.useful_info[0][0].name))
                demo_actions.append(
                    "put {} in {}".format(
                        self.useful_info[0][0].name, self.answer_box.name
                    )
                )
                return demo_actions
            elif self.useful_info[0][1] < self.useful_info[1][1]:
                # right is heavier
                demo_actions.append("take {}".format(self.useful_info[1][0].name))
                demo_actions.append(
                    "put {} in {}".format(
                        self.useful_info[1][0].name, self.answer_box.name
                    )
                )
                return demo_actions
            else:
                demo_actions.append("take {}".format(self.useful_info[0][0].name))
                demo_actions.append(
                    "put {} in {}".format(
                        self.useful_info[0][0].name, self.answer_box.name
                    )
                )

                demo_actions.append("take {}".format(self.useful_info[1][0].name))
                demo_actions.append(
                    "put {} in {}".format(
                        self.useful_info[1][0].name, self.answer_box.name
                    )
                )
                return demo_actions

        on_scale = [0, 1]
        for i in range(2, len(self.useful_info)):
            if self.useful_info[on_scale[0]][1] > self.useful_info[on_scale[1]][1]:
                demo_actions += [
                    "take {}".format(self.useful_info[on_scale[1]][0].name),
                    "take {}".format(self.useful_info[i][0].name),
                    "put {} in right side of the balance scale (ID: 4)".format(
                        self.useful_info[i][0].name
                    ),
                    "look",
                ]
                on_scale[1] = i
            else:
                demo_actions += [
                    "take {}".format(self.useful_info[on_scale[0]][0].name),
                    "take {}".format(self.useful_info[i][0].name),
                    "put {} in left side of the balance scale (ID: 3)".format(
                        self.useful_info[i][0].name
                    ),
                    "look",
                ]
                on_scale[0] = i
        max_weight = 0
        if self.useful_info[on_scale[0]][1] > self.useful_info[on_scale[1]][1]:
            demo_actions.append("take {}".format(self.useful_info[on_scale[0]][0].name))
            demo_actions.append(
                "put {} in {}".format(
                    self.useful_info[on_scale[0]][0].name, self.answer_box.name
                )
            )
            max_weight = self.useful_info[on_scale[0]][1]
        else:
            demo_actions.append("take {}".format(self.useful_info[on_scale[1]][0].name))
            demo_actions.append(
                "put {} in {}".format(
                    self.useful_info[on_scale[1]][0].name, self.answer_box.name
                )
            )
            max_weight = self.useful_info[on_scale[1]][1]

        for cube, cube_mass in self.useful_info:
            if cube_mass == max_weight:
                demo_actions += [
                    "take {}".format(cube.name),
                    "put {} in {}".format(cube.name, self.answer_box.name),
                ]

        return demo_actions
\end{lstlisting}
\begin{lstlisting}[breaklines = true, caption = {take-photo}]
    def get_demo_actions(self):
        speed = self.camera.properties["current_shutter_speed"]
        iso = self.camera.properties["current_iso"]
        aperture = self.camera.properties["current_aperture"]

        target_aperture = self.useful_info[0]
        target_speed = self.useful_info[1]
        target_iso = self.useful_info[2]

        adjust_actions = [
            "focus {}".format(self.target_food.name),
        ]
        if target_aperture > aperture:
            adjust_actions += ["rotate aperture clockwise"] * (
                target_aperture - aperture
            )
        elif target_aperture < aperture:
            adjust_actions += ["rotate aperture anticlockwise"] * (
                -target_aperture + aperture
            )
        else:
            pass
        if target_speed > speed:
            adjust_actions += ["rotate shutter speed clockwise"] * (
                target_speed - speed
            )
        elif target_speed < speed:
            adjust_actions += ["rotate shutter speed anticlockwise"] * (
                -target_speed + speed
            )
        else:
            pass
        if target_iso > iso:
            adjust_actions += ["rotate iso clockwise"] * (target_iso - iso)
        elif target_iso < iso:
            adjust_actions += ["rotate iso anticlockwise"] * (-target_iso + iso)
        else:
            pass

        demo_actions = ["take camera (ID: 2)"] + adjust_actions + ["press shutter"]

        return demo_actions
\end{lstlisting}
\begin{lstlisting}[breaklines = true, caption = {plant-tree}]
    def get_demo_actions(self):
        return [
            "take shovel (ID: 2)",
            "dig with shovel (ID: 2)",
            "take tree (ID: 8)",
            "look",
            "put tree (ID: 8) in hole (ID: 9)",
            "inventory",
            "put soil (ID: 10) in hole (ID: 9)",
            "take jug (ID: 7)",
            "put jug (ID: 7) in sink (ID: 5)",
            "turn on sink (ID: 5)",
            "turn off sink (ID: 5)",
            "take jug (ID: 7)",
            "pour water in jug (ID: 7) into soil (ID: 10)",
        ]
\end{lstlisting}
\begin{lstlisting}[breaklines = true, caption = {boil-water}]
    def get_demo_actions(self):
        return [
            "take pot (ID: 4)",
            "put pot (ID: 4) in sink (ID: 3)",
            "examine sink (ID: 3)",
            "turn on sink (ID: 3)",
            "examine sink (ID: 3)",
            "turn off sink (ID: 3)",
            "take pot (ID: 4)",
            "look",
            "put pot (ID: 4) on stove (ID: 2)",
            "examine stove (ID: 2)",
            "turn on stove (ID: 2)",
            "examine stove (ID: 2)",
            "examine stove (ID: 2)",
            "examine stove (ID: 2)",
        ]
\end{lstlisting}
\begin{lstlisting}[breaklines = true, caption = {forge-key}]
    def get_demo_actions(self):
        demo_actions = [
            "take copper ingot (ID: 4)",
            "put copper ingot (ID: 4) in foundry (ID: 3)",
            "turn on foundry (ID: 3)",
            "look",
            "look",
            "look",
            "look",
            "look",
            "look",
            "pour copper (liquid) (ID: 4) into key mold (ID: 6)",
            "look",
            "look",
            "take copper key (ID: 4)",
            "open door (ID: 5) with copper key (ID: 4)",
        ]

        return demo_actions
\end{lstlisting}
\begin{lstlisting}[breaklines = true, caption = {inclined-plane}]
    def get_demo_actions(self):
        look_table = {
            0.5: ["look"] * 5,
            1: ["look"] * 5,
            1.5: ["look"] * 5,
            2: ["look"] * 5,
        }

        a1 = self.useful_info[0][0]
        a2 = self.useful_info[0][1]
        a1_look = look_table[a1]
        a2_look = look_table[a2]

        demo_actions = (
            [
                "take stopwatch (ID: 5)",
                "take block (ID: 4)",
                "put block (ID: 4) on inclined plane 1 (ID: 2)",
                "activate stopwatch (ID: 5)",
            ]
            + a1_look
            + [
                "deactivate stopwatch (ID: 5)",
                "examine stopwatch (ID: 5)",
                "reset stopwatch (ID: 5)",
                "take block (ID: 4)",
                "put block (ID: 4) on inclined plane 2 (ID: 3)",
                "activate stopwatch (ID: 5)",
            ]
            + a2_look
            + [
                "deactivate stopwatch (ID: 5)",
                "examine stopwatch (ID: 5)",
            ]
        )
        if a1 > a2:
            return demo_actions + ["focus on inclined plane 2 (ID: 3)"]
        else:
            return demo_actions + ["focus on inclined plane 1 (ID: 2)"]
\end{lstlisting}
\begin{lstlisting}[breaklines = true, caption = {wash-clothes}]
    def get_demo_actions(self):
        washing, drying, busketing = [], [], []

        for cloth in self.dirty_clothes:
            washing += [
                "take {}".format(cloth.name),
                "put {} in washing machine (ID: 2)".format(cloth.name),
            ]

            drying += [
                "take {}".format(cloth.name),
                "put {} in dryer (ID: 3)".format(cloth.name),
            ]

            busketing += [
                "take {}".format(cloth.name),
                "put {} in basket (ID: 12)".format(cloth.name),
            ]

        for cloth in self.clean_clothes:
            busketing += [
                "take {}".format(cloth.name),
                "put {} in basket (ID: 12)".format(cloth.name),
            ]

        demo_actions = (
            [
                "open washing machine (ID: 2)",
            ]
            + washing
            + [
                "use bottle of detergent (ID: 4) on washing machine (ID: 2)",
                "close washing machine (ID: 2)",
                "turn on washing machine (ID: 2)",
                "wait",
                "look",
                "look",
                "open washing machine (ID: 2)",
                "open dryer (ID: 3)",
            ]
            + drying
            + [
                "close dryer (ID: 3)",
                "turn on dryer (ID: 3)",
                "wait",
                "look",
                "look",
                "open dryer (ID: 3)",
                "take skirt (ID: 5)",
            ]
            + busketing
        )

        return demo_actions
\end{lstlisting}

\subsection{Analysis of Demo Actions}
Figure~\ref{fig:steps} displays the numbers of steps of the generated rule-based policies for environments to complete the tasks. We note that the number of steps may vary due to the randomness in the environments. For example, in mix paint, if the target color is black, 3 steps are needed, and other colors may only require 2 steps. 
\begin{figure}[ht]
    \centering
    \includegraphics[width=\linewidth]{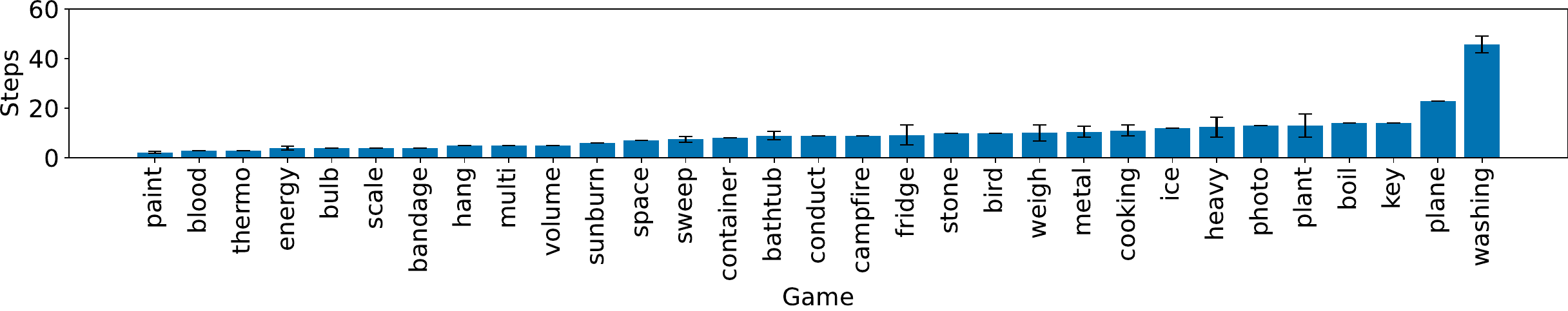}
    \caption{Steps to complete the tasks}
    \label{fig:steps}
\end{figure}



\clearpage
\section{Prompts for World Model}

\subsection{Prompt for Next State and Reward/Terminal Predictions}
\label{app:next_state_and_reward_prediction}
\begin{lstlisting}[breaklines=true,caption={Code for Prompts of Generating Potential Actions.}]
prompt = (
            "You are a simulator of a text game. Read the task description of a text game. "
            "Given the current game state in JSON, "
            "you need to decide the new game state after taking an action including the game score.\n"
        )
        prompt += (
            "Your response should be in the JSON format. "
            "It should have three keys: 'modified', 'removed', and 'score'. "
            "The 'modified' key stores a list of all the object states that are added or changed after taking the action. "
            "Keep it an empty list if no object is added or modified. "
            "The 'removed' key stores a list of uuids of the objects that are removed. "
            "Keep it an empty list if no object is removed. "
            "The 'score' key stores a JSON with three keys: "
            "'score', 'gameOver', and 'gameWon'. "
            "'score' stores the current game score, "
            "'gameOver' stores a bool value on whether the game is over, "
            "and 'gameWon' stores a bool value on whether the game is won. \n"
        )

        last_action = "" if len(self.last_actions) == 0 else self.last_actions[-1]
        max_UUID = importlib.import_module(self.game_name).UUID
        if current_state is None:
            current_state = get_state(self.game, last_action, max_UUID, self.game_name)
            current_state_for_prompt = make_game_state(current_state)
            max_uuid = current_state["max_UUID"]
        else:
            # print("use the predicted state")

            current_state_for_prompt = current_state
            max_uuid = len(current_state["game_state"])

            # start adding examples
        example_prompt = self.build_examples()
        prompt += example_prompt
        # end of adding examples
        # Task
        prompt += "Here is the game that you need to simulate:\n"
        prompt += "Task Description:\n"
        prompt += f"{self.task_desc}\n"

        # load rules
        obj_desc = preprocess_obj_desc(self.obj_rules[self.game_name])
        action_desc = self.action_rules[self.game_name]
        score_desc = self.score_rules[self.game_name]

        prompt += "Here are the descriptions of all game objects properties:\n"
        prompt += obj_desc.strip()
        prompt += "\n"
        prompt += "Here are the descriptions of all game actions:\n"
        prompt += action_desc.strip()
        prompt += "\n"
        prompt += "Here is a description of the game score function:\n"
        prompt += score_desc.strip()
        prompt += "\n"

        # data_state, data_UUID_base, data_action = None, None, None
        prompt += "Here is the game state:\n"
        prompt += f"{current_state_for_prompt}\n"
        prompt += "\n"

        prompt += f"The current game UUID base is {max_uuid}\n"
        prompt += f"The action to take is:\n{action}\n"
\end{lstlisting}

\begin{lstlisting}[breaklines=true,caption={{The Predicted Next State and Reward/Terminal in lit-lightbulb (bulb) environment.}}]
{'game_state': [{'name': 'room (ID: 1)', 'uuid': 1, 'type': 'World', 'properties': {'isContainer': True, 'isMoveable': True, 'isOpenable': False, 'isOpen': True, 'containerPrefix': 'in'}, 'contains': ['agent (ID: 0)', 'light bulb (ID: 2)', 'red wire (ID: 3)', 'black wire (ID: 4)', 'blue wire (ID: 5)', 'battery (ID: 6)']}, {'name': 'agent (ID: 0)', 'uuid': 0, 'type': 'Agent', 'properties': {'isContainer': True, 'isMoveable': True, 'isOpenable': False, 'isOpen': True, 'containerPrefix': 'in'}, 'contains': []}, {'name': 'light bulb (ID: 2)', 'uuid': 2, 'type': 'LightBulb', 'properties': {'isContainer': False, 'isMoveable': True, 'is_electrical_object': True, 'conductive': True, 'connects': {'terminal1': [3, 'terminal1'], 'terminal2': [None, None]}, 'on': False}, 'contains': []}, {'name': 'red wire (ID: 3)', 'uuid': 3, 'type': 'Wire', 'properties': {'isContainer': False, 'isMoveable': True, 'is_electrical_object': True, 'conductive': True, 'connects': {'terminal1': [2, 'terminal1'], 'terminal2': [None, None]}, 'is_wire': True}, 'contains': []}, {'name': 'black wire (ID: 4)', 'uuid': 4, 'type': 'Wire', 'properties': {'isContainer': False, 'isMoveable': True, 'is_electrical_object': True, 'conductive': True, 'connects': {'terminal1': (None, None), 'terminal2': (None, None)}, 'is_wire': True}, 'contains': []}, {'name': 'blue wire (ID: 5)', 'uuid': 5, 'type': 'Wire', 'properties': {'isContainer': False, 'isMoveable': True, 'is_electrical_object': True, 'conductive': True, 'connects': {'terminal1': (None, None), 'terminal2': (None, None)}, 'is_wire': True}, 'contains': []}, {'name': 'battery (ID: 6)', 'uuid': 6, 'type': 'Battery', 'properties': {'isContainer': False, 'isMoveable': True, 'is_electrical_object': True, 'conductive': True, 'connects': {'cathode': (None, None), 'anode': (None, None)}}, 'contains': []}]}, {'score': 0, 'gameOver': False, 'gameWon': False}
\end{lstlisting}
\subsection{Prompts of Generating Potential Actions.}
\label{app:prompt_action_proposal}

\begin{lstlisting}[breaklines=true,caption={Code for Prompts of Generating Potential Actions.}]
        prompt = (
            "You are a simulator of a text game. "
            "Read the task description and the descriptions of all game actions of a text game. "
            "Given the current game state in JSON, and the previous actions that lead to the current game state, "
            "you need to decide the most {} actions "
            "that can help to complete the task step by step at the current state.\n".format(
                k
            )
        )
        prompt += (
            "Each of your action should in one phrase with one verb and the objects it operates on. "
            "Examples of actions includes:\n"
            "move south" + ",\n"
            "detect with metal detector (ID: 15)" + ",\n"
            "dig with shovel (ID: 16)" + ",\n"
            "open freezer (ID: 2)" + ",\n"
            "put ice cube tray (ID: 3) in sink (ID: 4)" + ",\n"
            "dice patato (ID: 2) with knife (ID: 8)" + ",\n"
            "give Type O negative blood (ID: 3) to patient (ID: 2)" + ",\n"
            "read cook book (ID: 7)" + ".\n"
        )

        prompt += (
            "Your response should be in the JSON format. "
            "It should have one key: 'avail_actions', which includes the list of the recommended actions. \n"
        )

        last_action = "" if len(self.last_actions) == 0 else self.last_actions[-1]
        max_UUID = importlib.import_module(self.game_name).UUID
        if current_state is None:
            current_state = get_state(self.game, last_action, max_UUID, self.game_name)
            current_state_for_prompt = make_game_state(current_state)
            max_uuid = current_state["max_UUID"]
        else:
            # print("use the predicted state")

            current_state_for_prompt = current_state
            max_uuid = len(current_state["game_state"])

            # start adding examples
        # example_prompt = self.build_examples()
        # prompt += example_prompt
        # end of adding examples
        # Task
        prompt += "Here is the game that you need to simulate:\n"
        prompt += "Task Description:\n"
        prompt += f"{self.task_desc}\n"

        # load rules
        obj_desc = preprocess_obj_desc(self.obj_rules[self.game_name])
        action_desc = self.action_rules[self.game_name]
        score_desc = self.score_rules[self.game_name]

        prompt += "Here are the descriptions of all game objects properties:\n"
        prompt += obj_desc.strip()
        prompt += "\n"
        prompt += "Here are the descriptions of all game actions:\n"
        prompt += action_desc.strip()
        prompt += "\n"
        prompt += "Here is a description of the game score function:\n"
        prompt += score_desc.strip()
        prompt += "\n"

        # data_state, data_UUID_base, data_action = None, None, None
        prompt += "Here is the game state:\n"
        prompt += f"{current_state_for_prompt}\n"
        prompt += "\n"

        prompt += f"The current game UUID base is {max_uuid}\n"

        if len(self.last_actions) == 0:
            prompt += "There is no previous actions."
        else:
            prompt += "The previous actions {}:\n".format(
                "is" if len(self.last_actions) == 1 else "are"
            )
            for action in self.last_actions:
                prompt += action + "\n"
\end{lstlisting}

\begin{lstlisting}[breaklines=true,caption={{The Generated Actions of make-ice-cubes (ice) environment}.}]
{'avail_actions': ['open freezer (ID: 2)', 'take ice cube tray (ID: 3) from freezer (ID: 2)', 'put ice cube tray (ID: 3) in sink (ID: 4)', 'turn on sink (ID: 4)', 'take ice cube tray (ID: 3) from sink (ID: 4)', 'put ice cube tray (ID: 3) in freezer (ID: 2)', 'close freezer (ID: 2)', 'wait for ice to form', 'open freezer (ID: 2)', 'check ice cube tray (ID: 3) for ice']}
\end{lstlisting}

\clearpage
\section{Action Matching}
{We provide a detailed explanation of the action matching process:
\begin{itemize}[leftmargin=*, topsep=0pt, partopsep=0pt, parsep=0pt, itemsep=0pt]
\item Embedding model: We use OpenAI's \texttt{text-embedding-3-small} for the matching of actions. 
\item Candidate set $A'$ formation: At each state $s_t$, the game engine exposes the full set of valid actions through its \texttt{getValidActions()} interface. This set typically contains 500–800 actions depending on the environment and the current state. We enumerate all valid actions returned by the environment at the current state to form $A'$. We have added this clarification to the revised manuscript.
\item Threshold and tie-breaking: We do not apply any similarity threshold. we always select the action with the highest cosine similarity. In cases of exact ties (which are extremely rare in practice due to the continuous nature of embedding vectors), we select the first action in the enumeration order. We have added this specification to the revised paper.
\end{itemize}}

\section{Format Error Handling}
{As noted in the main context, we observe that format errors occasionally occur during evaluation. To mitigate this, we rerun the experiments to ensure robustness. Since we primarily evaluate closed-source models, our ability to intervene at the decoding or post-processing level is inherently limited. Nevertheless, these occasional format errors do not affect our overall conclusions. Moreover, as large language models continue to evolve, their instruction-following capabilities are expected to improve steadily, and such format errors should become increasingly rare in future iterations.}

\clearpage
\section{Accuracy of Policy Verification}
\label{app:verify}

We provide the accuracy of the policy verification regarding the three criteria, i.e., score, gameWon and gameOver. We note that the performance on gameWon and gameOver predictions are far better than the prediction of score.

\begin{figure}[ht]
\centering
    \includegraphics[width=\columnwidth]{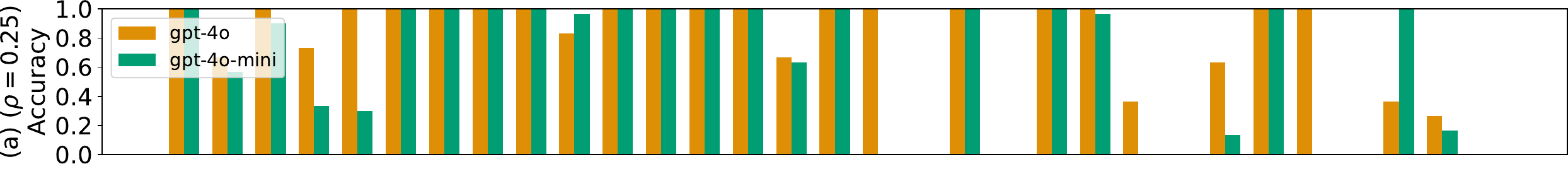}
    \includegraphics[width=\columnwidth]{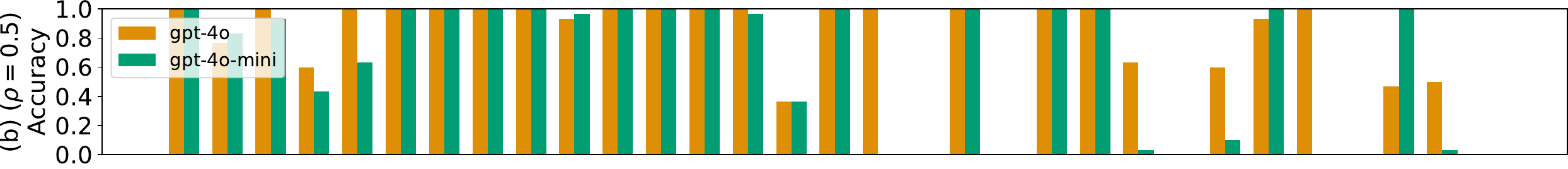}
    \includegraphics[width=\columnwidth]{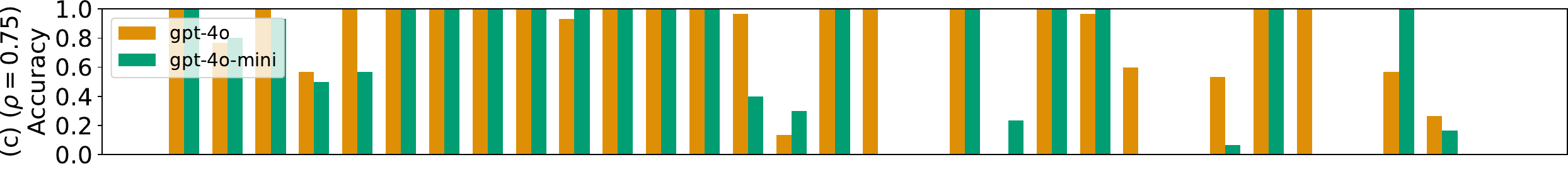}

    \includegraphics[width=\columnwidth]{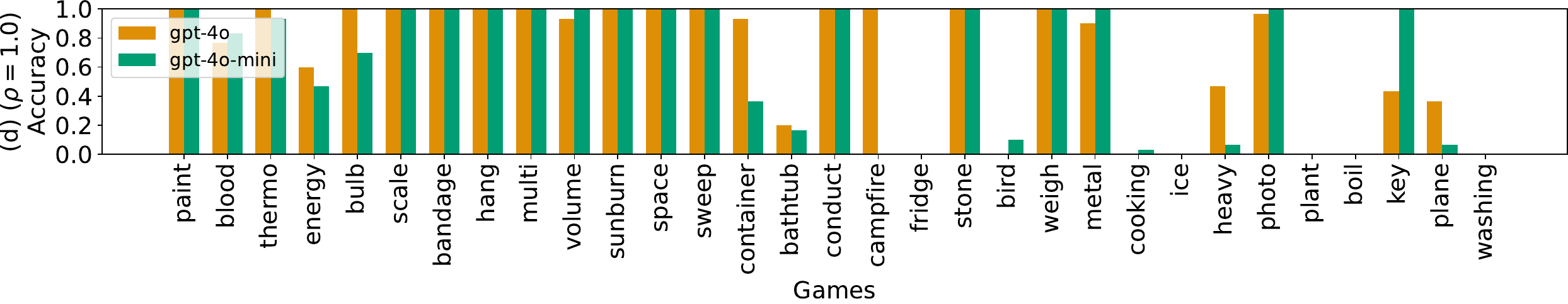}
     
\caption{The accuracy of the world model to verify the correct policies on score}
\label{fig:verify_score}
\end{figure}

\begin{figure}[ht]
\centering
    \includegraphics[width=\columnwidth]{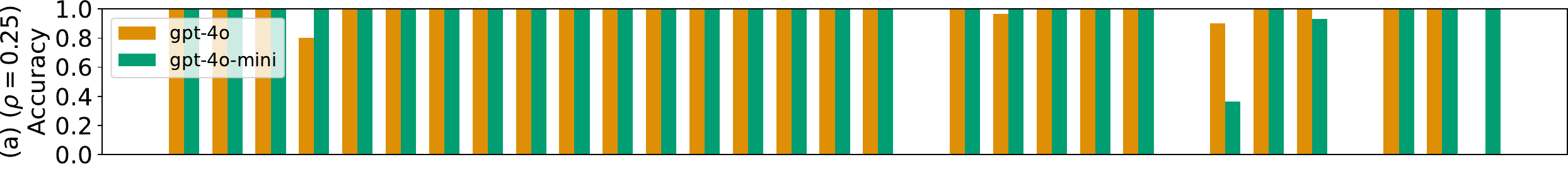}
    \includegraphics[width=\columnwidth]{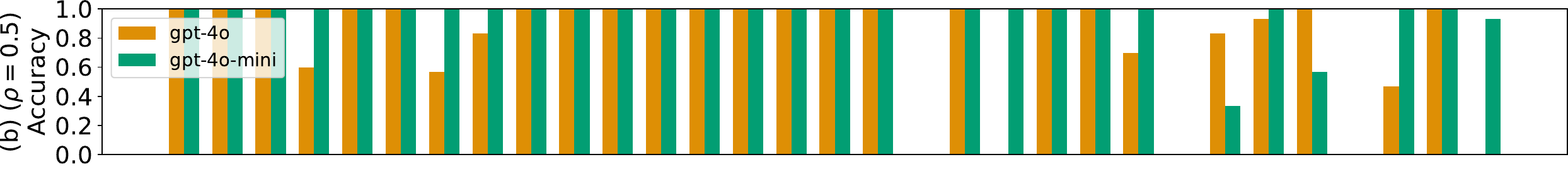}
    \includegraphics[width=\columnwidth]{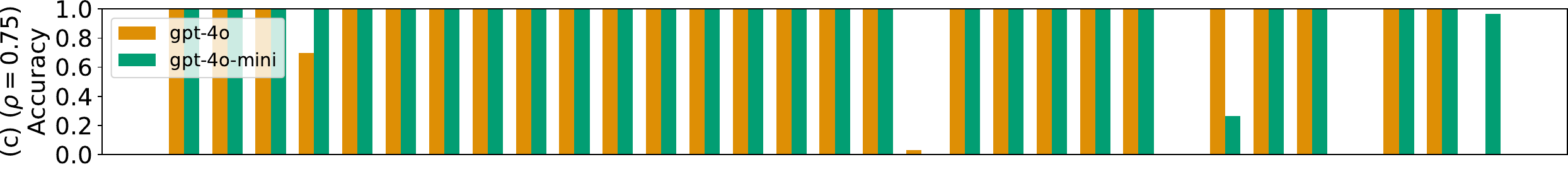}

    \includegraphics[width=\columnwidth]{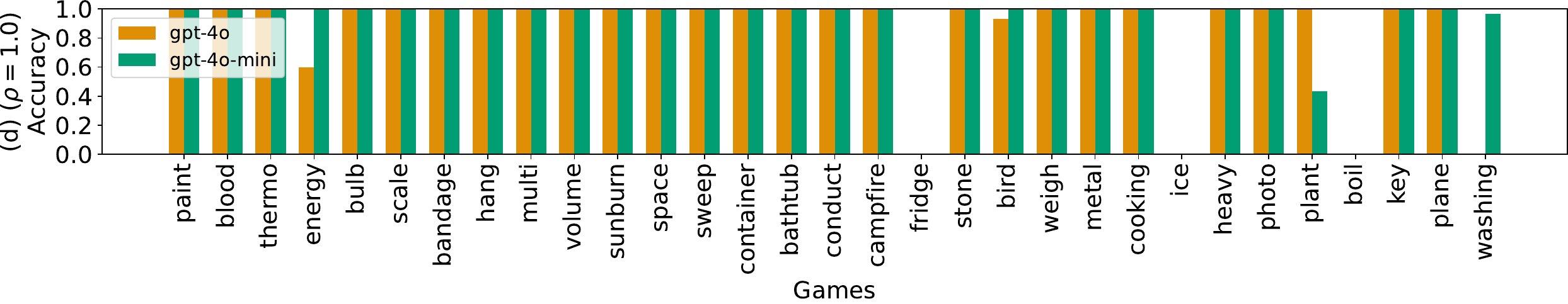}
     
\caption{The accuracy of the world model to verify the correct policies on gameOver}
\label{fig:verify_gameOver}
\end{figure}

\begin{figure}[ht]
\centering
    \includegraphics[width=\columnwidth]{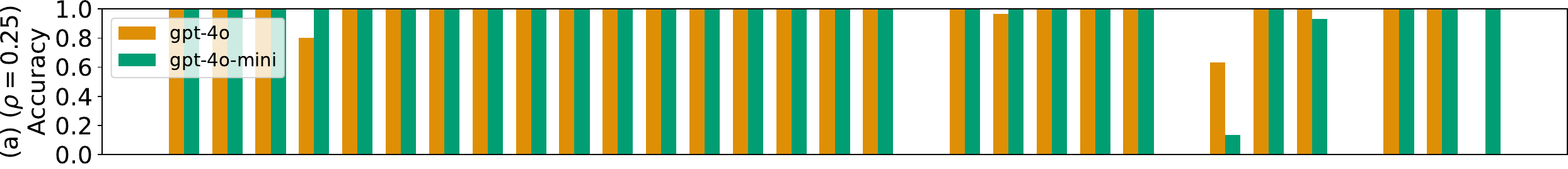}
    \includegraphics[width=\columnwidth]{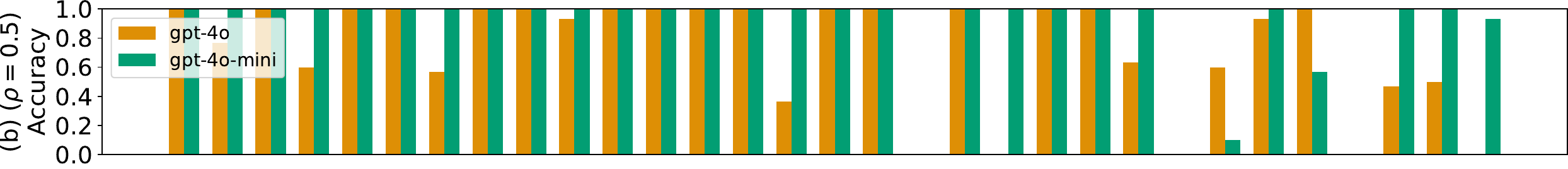}
    \includegraphics[width=\columnwidth]{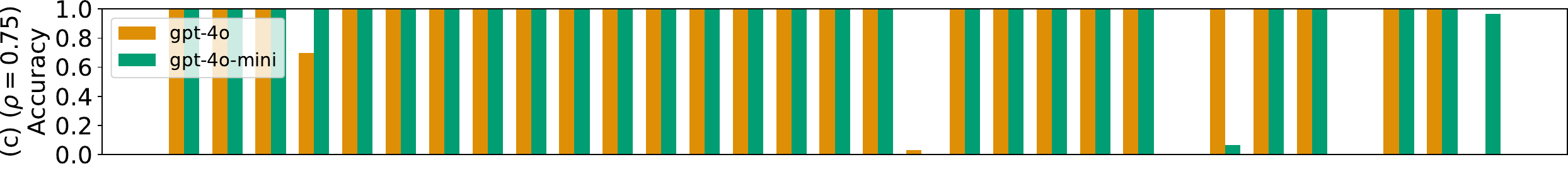}

    \includegraphics[width=\columnwidth]{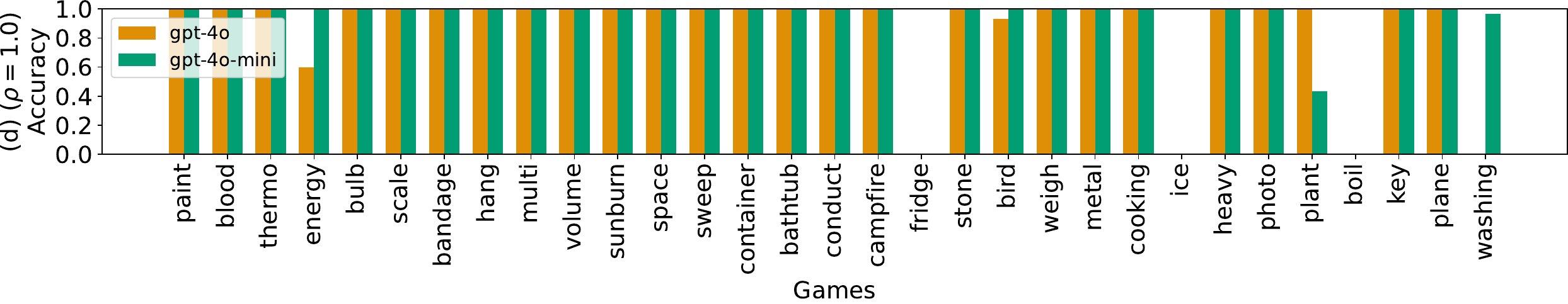}
     
\caption{The accuracy of the world model to verify the correct policies on gameWon}
\label{fig:verify_gameWon}
\end{figure}

\clearpage
\section{Step Accuracy of Action Proposal}
\label{app:step_accuracy}
We also provide the accuracy of each steps for the action proposal tasks. We observe for most of the task, there is some key steps that the world model has low accuracy for the action proposal, which brings difficulties for the world model to complete the tasks. 
\begin{figure}[ht]
    \centering
    \includegraphics[width=0.75\linewidth]{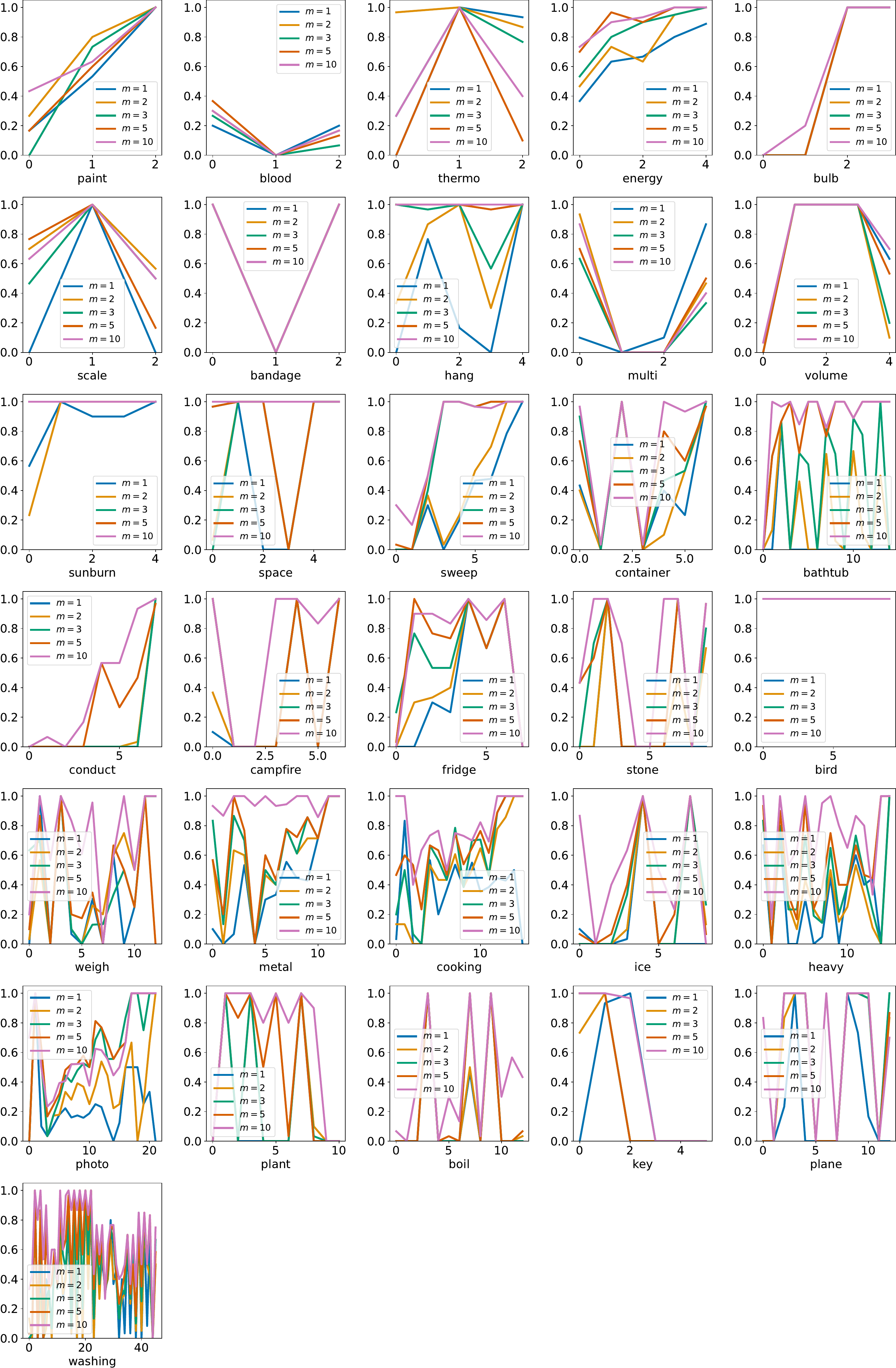}
    \caption{Step correctness of the action proposal of GPT-4o-mini}
    \label{fig:step_gpt4o_mini}
\end{figure}

\begin{figure}
    \centering
    \includegraphics[width=0.75\linewidth]{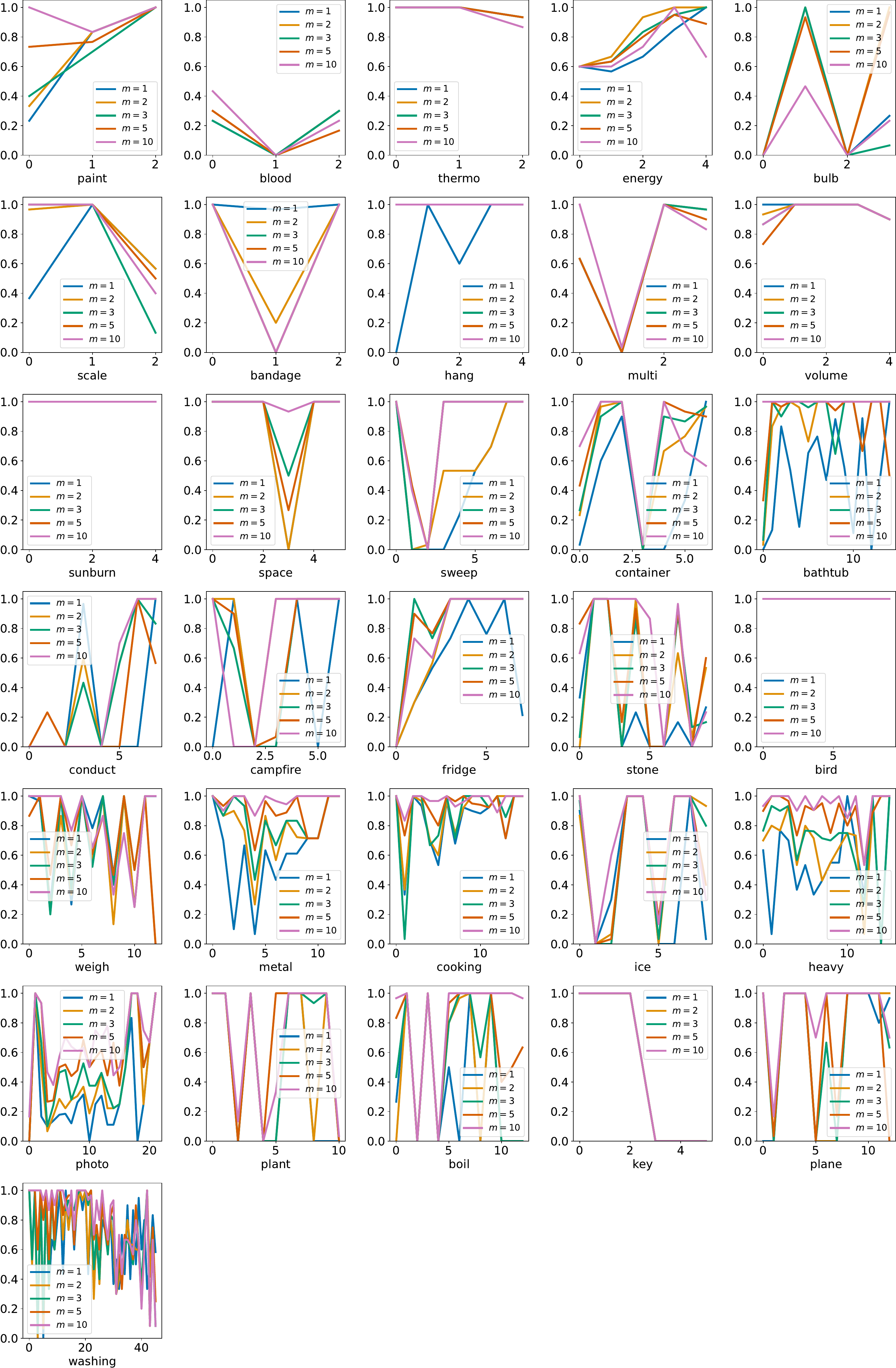}
    \caption{Step correctness of the action proposal of GPT-4o}
    \label{fig:step_gpt4o}
\end{figure}

\end{document}